\documentclass[10pt,twocolumn,letterpaper]{article}

\usepackage[pagenumbers]{cvpr} 

\usepackage[dvipsnames]{xcolor}

\usepackage{graphicx}
\usepackage{amsmath}
\usepackage{amssymb}
\usepackage{booktabs}\usepackage{algorithm}
\usepackage{url} 
\usepackage{float}
\usepackage{color}
\usepackage{bm}
\usepackage{bbm}
\usepackage{algorithm}
\usepackage{algpseudocode}
\usepackage{enumitem}
\usepackage{siunitx}
\usepackage{multirow}
\usepackage{amsfonts}       
\usepackage{nicefrac}       
\usepackage{microtype}      
\usepackage{xcolor}         
\usepackage{diagbox} 
\usepackage{tabularx}
\usepackage{hhline}
\usepackage{arydshln}		
\usepackage{mathtools} 
\usepackage{amsthm, soul}			
\usepackage{makecell}

\def\defn{\,\coloneqq\,}
\def\argmin{\mathop{\mathrm{arg\,min}}} 

\def\cbm{{\bm{c}}}
\def\ebm{{\bm{e}}}

\def\xbm{{\bm{x}}}
\def\gbm{{\bm{g}}}
\def\ybm{{\bm{y}}}

\def\kbm{{\bm{k}}}
\def\hbm{{\bm{h}}}
\def\fbm{{\bm{f}}}
\def\obm{{\bm{o}}}

\def\thetabm{{\bm{\theta}}}
\def\phibm{{\bm{\phi}}}
\def\zerobm{\bm{0}}

\def\Xbm{{\bm{X}}}

\def\epsilonbm{{\bm{\epsilon}}}
\def\Ibf{{\mathbf{A}}}
\def\Ibf{{\mathbf{B}}}
\def\Ibf{{\mathbf{C}}}
\def\Ibf{{\mathbf{D}}}
\def\Ibf{{\mathbf{E}}}
\def\Ibf{{\mathbf{F}}}
\def\Ibf{{\mathbf{G}}}
\def\Ibf{{\mathbf{H}}}
\def\Ibf{{\mathbf{I}}}

\def\Tsf{{\mathsf{T}}}

\def\R{\mathbb{R}}
\def\E{\mathbb{E}}

\def\Hcal{{\mathcal{H}}}

\def\Ncal{{\mathcal{N}}}

\def\Acal{{\mathcal{A}}}
\def\Lcal{{\mathcal{L}}}
\def\Dcal{{\mathcal{D}}}

\def\Ecal{{\mathcal{E}}}
\definecolor{lightgreen}{rgb}{.9,1,.9}
\definecolor{lightblue}{rgb}{.9,.9,1.}

\definecolor{cvprblue}{rgb}{0.21,0.49,0.74}
\usepackage[pagebackref,breaklinks,colorlinks,citecolor=cvprblue]{hyperref}

\def\paperID{550} 
\def\confName{CVPR}
\def\confYear{2024}

\title{FLAIR: A Conditional Diffusion Framework with Applications \\ to Face Video Restoration}

\author{Zihao Zou$^{\footnotesize *}$, ~~Jiaming Liu$^{\footnotesize *}$,  ~~Shirin Shoushtari, ~~Yubo Wang\\ Weijie Gan, ~~and ~~Ulugbek S. Kamilov\\
Washington University in St.~Louis, MO, USA\\
$^{\footnotesize *}$\small These authors contributed equally.\\
}

\begin{document}
\maketitle

\begin{abstract}
Face video restoration (FVR) is a challenging but important problem where one seeks to recover a perceptually realistic face videos from a low-quality input. While diffusion probabilistic models (DPMs) have been shown to achieve remarkable performance for face image restoration, they often fail to preserve temporally coherent, high-quality videos, compromising the fidelity of reconstructed faces. We present a new conditional diffusion framework called FLAIR for FVR. FLAIR ensures temporal consistency across frames in a computationally efficient fashion by converting a traditional image DPM into a video DPM. The proposed conversion uses a recurrent video refinement layer and a temporal self-attention at different scales. FLAIR also uses a conditional iterative refinement process to balance the perceptual and distortion quality during inference. This process consists of two key components: a data-consistency module that analytically ensures that the generated video precisely matches its degraded observation and a coarse-to-fine image enhancement module specifically for facial regions. Our extensive experiments show superiority of FLAIR over the current state-of-the-art (SOTA) for video super-resolution, deblurring, JPEG restoration, and space-time frame interpolation on two high-quality face video datasets. 
\end{abstract}

\section{Introduction}
\label{sec:intro}

\begin{figure*}[t!]
	\centering	\includegraphics[width=0.95\textwidth]{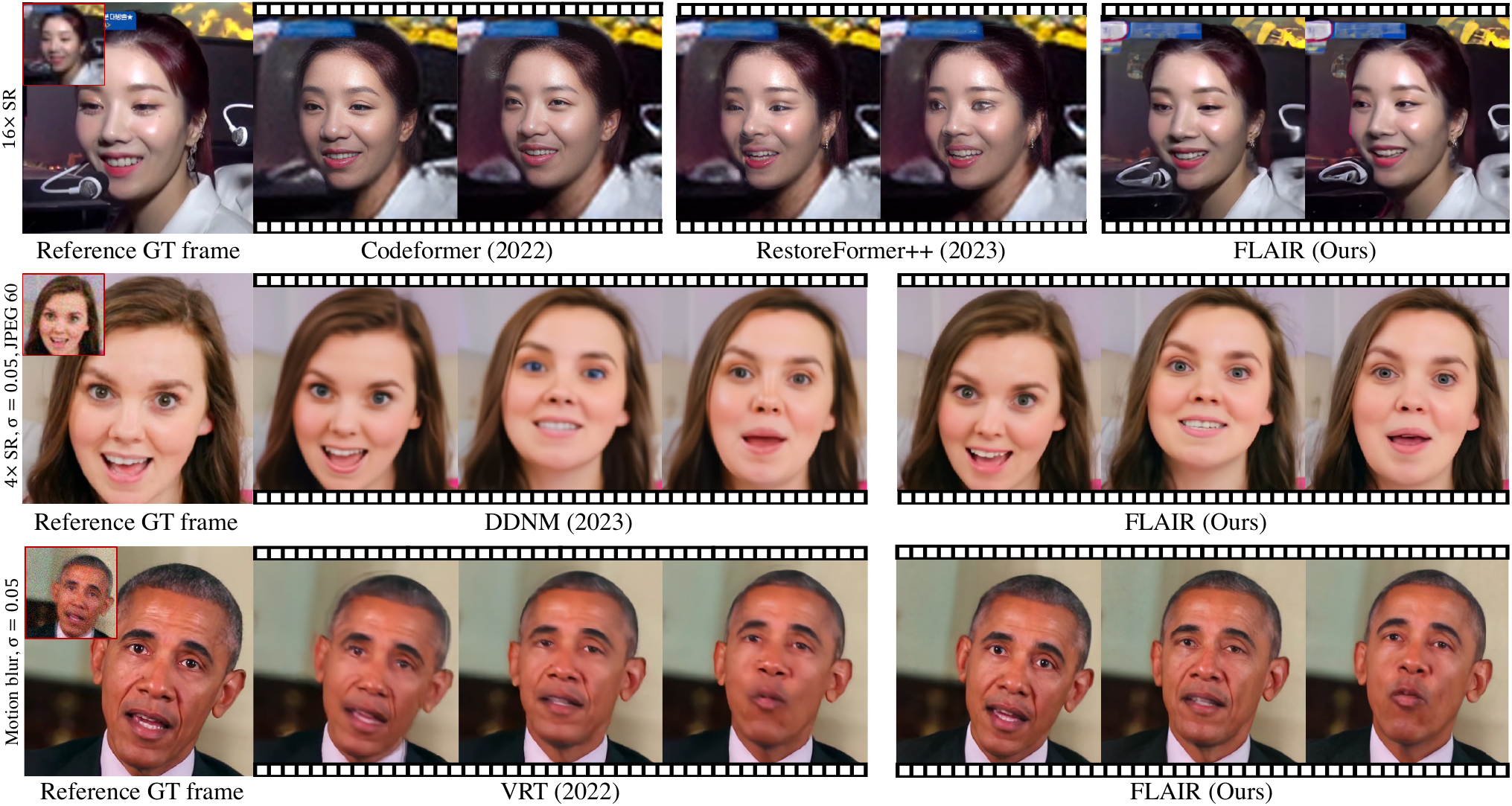}
    \vspace{-0.5em}
	\caption{Qualitative evaluation of the proposed FLAIR method.~\emph{\textbf{Top:}} FLAIR can restore high-quality facial details and preserve the data fidelity across frames, while both Codeformer~\cite{Zhou.etal2022towards} and RestoreFormer++~\cite{Wang2023.etalrestoreformer++} hallucinate faces that diverge from the original subject.~\emph{\textbf{Middle:}} FLAIR produces better temporal consistency than existing conditional diffusion method DDNM~\cite{Wang.etal2023zeroshot}.~\emph{\textbf{Bottom:}} FLAIR preserves more high-frequency details for motion deblurring, delivering superior perceptual quality than video restoration method VRT~\cite{Liang.etal2022vrt}.}
	\label{Fig:fig0}
\end{figure*}

As a subcategory of the general image and video restoration~\cite{Tu.etal2022maxim, zhang.etal2020a, Liang2021swinir, Liang2022recurrent}, face restoration is an active research area in computer vision~\cite{Roth.etal2016, Li.etal2017generative, Tzimiropoulos.etal2015project, Jourabloo.2017etal, Kumar.etal2020, Li_2022_CVPR}. Image and video restoration is usually~\emph{ill-posed} due to the information loss induced by degradation~(\emph{e.g.,} resolution loss, blur, encoding artifacts, and noise), with multiple plausible high-quality (HQ) objects leading to the same low-quality (LQ) observation. Face restoration has recently been greatly improved by using generative priors~\cite{Yang.etal2021gan, He.etal2022gcfsr, Wang.etal2021towards} and pre-trained face dictionary priors ~\cite{Li.etal2020blind, Zhou.etal2022towards, Gu.etal2022vqfr, Wang.etal2022restoreformer}.  While SOTA methods---such as Codeformer~\cite{Zhou.etal2022towards}, VQFR~\cite{Gu.etal2022vqfr}, and RestoreFormer~\cite{Wang.etal2022restoreformer}---can restore high-quality results with fine details, they usually hallucinate HQ faces that diverge from the original subjects in the presence of severe degradation~\cite{Zhao.etal2023towards}, leading to large distortion, as can be seen in Fig.~\ref{Fig:fig0} (\emph{Top}).

Diffusion probabilistic models (DPMs)~\cite{Hao.etal2020, Song.etal2020b} have attracted significant attention as an alternative to traditional generative models due to their excellent performance in image and video generation~\cite{Dhariwal.etal2021, Saharia.etal2022b, Rombach.etal2022high, Zhang2023adding, Blattmann.etal2023align, Harvey2022flexible}. DPMs have been applied to a range of imaging problems, showing impressive results for face restoration. These methods generally fall into two categories: model-based unsupervised methods~\cite{Wang.etal2023dr2, Kawar.etal2022denoising, Laroche.etal2023fast, Chung.etal2022a, Wang.etal2023zeroshot, Song.etal2022pseudoinverse} and conditional training methods~\cite{Saharia.etal2022a, Saharia.etal2022palette, Whang.etal2022, Ren.etal2023multiscale}. Despite recent activity in the area, there are very few DPM-based frameworks for video restoration, especially in the context of face video restoration (FVR). The key challenges are the significant computational cost of training on video data and the lack of large-scale, publicly available HQ face video datasets. Given the stochasticity of the generative process in DPMs, another challenge is the effective use of nearby, similar but misaligned frames for reconstructing temporally aligned HQ reference frames~\cite{liu.etal2022video, Wang.etal2022survey}. For instance, as shown in Fig.~\ref{Fig:fig0}~(\emph{Middle}), one of the latest conditional image DPM, DDNM~\cite{Wang.etal2023zeroshot}, fails to produce a consistent facial restoration across frames.

\medskip\noindent
\textbf{Proposed Work:} We present \emph{Di\textbf{f}fusion Probabi\textbf{l}istic F\textbf{a}ce V\textbf{i}deo \textbf{R}estoration (FLAIR)}, a conditional generative model for FVR, that can generate multiple distinct, high-quality, enhanced face videos from a given degraded sequential data. We design FLAIR as a ``repeated-refinement'' conditional DPM. Instead of directly training on high-resolution videos, we first pre-train our conditional DPMs on images only, which allows us to use large-scale HQ image datasets very efficiently. The image DPMs are trained to take the degraded estimation as an auxiliary input for conditional restoration similar to~\cite{Saharia.etal2022a, Whang.etal2022, Liu.etal2023dolce}. Given a pre-trained image DPM backbone based on UNet~\cite{Dhariwal.etal2021},  we then modify it into a video restoration model by introducing a temporal dimension into the feature space of the neural network and only train these temporal layers on video sequences. Specifically, we propose a flow-guided video enhancement layer with a multi-scale recurrent module at the high-resolution scales of the UNet backbone, along with several temporal self-attention blocks that process the low resolution features in a sliding-window fashion. FLAIR is thus designed to capture long-range temporal dependencies, using information from multiple neighbouring frames for the restoration of each frame during inference. 

\begin{figure*}[t!]
	\centering
	\includegraphics[width=0.90\textwidth]{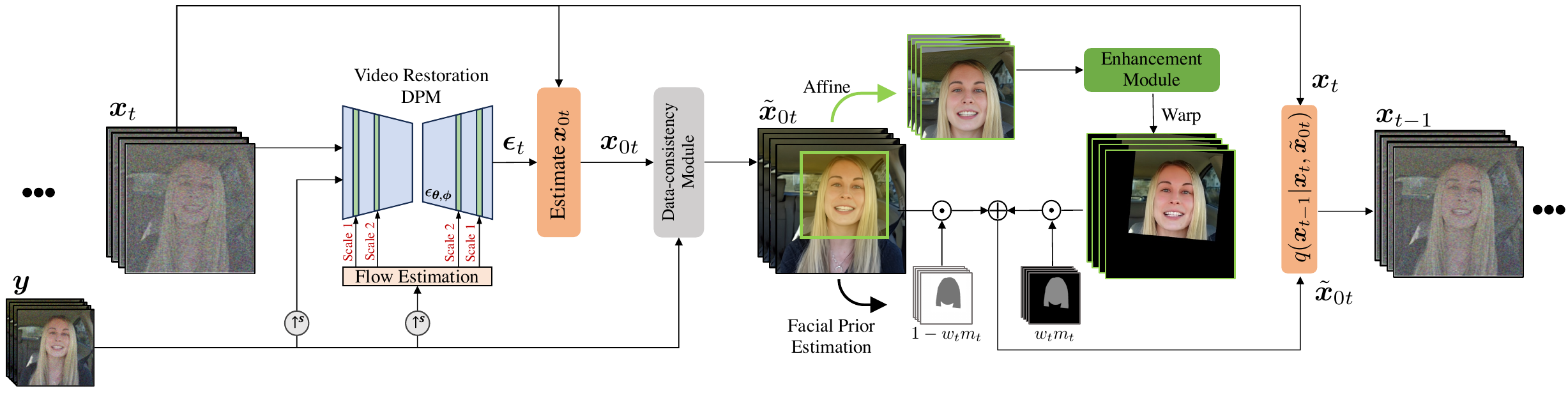}
    \vspace{-1em}
	\caption{Overview of the proposed FLAIR framework. At the $t$-th sampling step, FLAIR uses the degraded video frame $\ybm$ as the guidance for the video DPM to denoise the latent video sequence $\xbm_t$. The estimated $\xbm_{0t}$ is passed through the data-consistency module to ensure that its low-frequencies are consistent with $\ybm$. The enhancement module then improves faces from $\tilde\xbm_{0t}$ for next sampling step.} 
	\label{Fig:fig1}
\end{figure*}

To better balance the perceptual quality and data-fidelity~\cite{Blau.etal2018perception}, we propose a two-stage refinement process at every reverse diffusion step. The first stage involves an interpretable data-consistency (DC) module to analytically ensure that the generated coarse, clean intermediate results precisely match their LQ counterparts, even amid a range of mixed real-world degradations (\emph{e.g.,} a mix of resolution loss, blur, and JPEG). In the second stage, the DC outputs are further processed by a enhancement module for high-quality details specified for facial regions (see Fig.~\ref{Fig:fig1}). This design ensures that the enhancement module is compatible with various choices of restoration methods, enabling FLAIR to produce both perceptually realistic and data-consistent results.

Our main contributions can be summarized as follows: (1) We propose FLAIR as the first conditional diffusion framework for the recovery of long-term consistent, high-quality face videos from their LQ observations. Our key insight is to convert pre-trained image DPMs into video restoration models by inserting temporal layers that learn to align images in a temporally consistent manner (Fig.~\ref{Fig:fig2}). (2) Together with a data-consistency module and an enhancement module, we employ FLAIR in a two-stage conditional refinement process at each iteration of the reverse diffusion to further improve the perception and fidelity simultaneously. (3) We show through extensive experiments that FLAIR outperforms SOTA methods for composite noisy degradation on two high-quality face video datasets both quantitatively and qualitatively, showing great potential for practical applications.

\section{Related Work}

\label{sec:related}
\textbf{Face Restoration.} Traditional approaches for face restoration are based on the incorporation of prior knowledge and degradation models~\cite{Tang.etal2003, Chakrabarti.etal2007super, Gunturk.etal2003}. The quality of restored faces has been progressively improving after adoption of convolutional neural networks (CNNs)~\cite{Tuzel.etal2016global, Yu.etal2016ultra, Huang.etal2017wavelet, Zhang.etal2018super}. Recent work has investigated various deep priors for face image restoration, including geometric and reference priors~\cite{Chen.etal2018fsrnet, Bulat.etal2018super, Chen.etal2021semantic, Li.etal2018learning, Dogan.etal2019exemplar}. The restoration quality has been further improved by adapting pre-trained GANs, such as StyleGAN~\cite{karras.etal2019style}, as generative priors~\cite{Asim.etal2020blind, Yang.etal2020hifacegan, Wang.etal2021towards, Yang.etal2021gan, He.etal2022gcfsr}. This line of works treats face restoration as a conditional image generation problem by projecting the
LQ faces into a compact, low-dimension space of
the pre-trained generator. Another line of works,~\emph{e.g.,} VQFR~\cite{Gu.etal2022vqfr}, CodeFormer~\cite{Zhou.etal2022towards}, ResotreFormer~\cite{Wang.etal2022restoreformer} and its variant~\cite{Wang2023.etalrestoreformer++}, leverages pre-trained Vector-Quantization (VQ) codebooks~\cite{Esser.etal2021taming} as dictionaries learned on facial regions, achieving SOTA results in blind face restoration. 

\medskip\noindent
\textbf{Diffusion Models.} Denoising diffusion models~\cite{Hao.etal2020, Dhariwal.etal2021, Kingma.etal2021} and score-based models~\cite{Song.etal2019, Song.etal2020a, Song.etal2020b} are two related classes of generative models that were shown to achieve SOTA performance for unconditional image and video generation. Apart from unconditional image generation, diffusion models have been extensively investigated in various imaging restoration tasks. One line of works has focused on designing conditional training methods in a supervised fashion~\cite{Saharia.etal2022a, Whang.etal2022, Ren.etal2023multiscale, Delbracio.etal2023inversion, Saharia.etal2022palette}. Another line of work has focused on keeping the training of an unconditional image DPM intact, and only modify the inference procedure to enable sampling from a conditional distribution~\cite{Choi.etal2021, Chung.etal2022b, Chung.etal2022c, Kawar.etal2022denoising, Lugmayr.etal2022, Wang.etal2023zeroshot, Wang.etal2023dr2}.  However, only few DPMs methods~\cite{Zhou.etal2022cadm, Danier.etal2023ldmvfi, Chang2023.etallook} have been tried for image video enhancement and restoration. Notably, none of these methods have directly addressed video restoration tasks with a focus on FVR. 

\section{Preliminaries}
\label{sec:prelim}
\medskip\noindent
\textbf{Diffusion Probabilistic Models.} 
The forward process of DPMs~\cite{Hao.etal2020, Sohl.etal2015} is a Markov Chain that gradually adds noise $\epsilonbm\sim\Ncal(\zerobm,\Ibf)$ to data
$\xbm_0\sim q(\xbm_0), \xbm_0\in\R^d$ according to the variance schedule $\beta_t \in (0,1)$  for all $t = 1,\cdots,T$.
The Markov chain sequentially samples the noisy latent variables 
$\xbm_{1:T}$ with the same dimensionality as $\xbm_0$.
Using the notation $\alpha_t \defn 1- \beta_t$ and $\Bar{\alpha}_t \defn \Pi_{s=1}^t \alpha_s$, sampling of $\xbm_t$ given $\xbm_0$ can be expressed in a closed form 
\begin{equation}
    \label{eq:sampleNoisfwd}
    q(\xbm_t|\xbm_{0}) := \Ncal(\xbm_t; \sqrt{\Bar{\alpha}_t} \xbm_{0}, ( 1- \Bar{\alpha}_t) \Ibf).
\end{equation}
The unconditional generative reverse process is a Gaussian transition that samples from $\xbm_T\sim\Ncal(\zerobm,\Ibf)$ to $\xbm_0$ as
\begin{equation}
    \label{eq:sampleNoisbpd}
    q(\xbm_{t-1}|\xbm_t, \xbm_{0}) := \Ncal(\xbm_{t-1}; \mu_t(\xbm_t, \xbm_0), \sigma_{t}^2\Ibf),
\end{equation}
where $\mu_t(\xbm_t, \xbm_0)$ and $\sigma_{t}$ depend on $\xbm_t, \xbm_0$, and $\beta_t$. DPMs train $\epsilon_\thetabm$ to  learn the Gaussian transition $p_\thetabm(\xbm_{t-1}|\xbm_t)$ as an approximation of reverse diffusion  $q(\xbm_{t-1}|\xbm_t, \xbm_{0})$. By training the residual denoiser network $\epsilon_{\thetabm}(\xbm_t,t)$ to predict the total noise $\bm\epsilon_t$, one can estimate ${\xbm_{0t}}$ through
\begin{equation}
    \label{eq:x0t}
     \xbm_{0t}=(\xbm_t-\sqrt{1-\bar\alpha_t}\epsilon_{\thetabm}(\xbm_t,t))/\sqrt{\bar\alpha_t},
\end{equation}
where ${\xbm_{0t}}$ denotes the first prediction of $\xbm_0$ given the noisy observation $\xbm_t$. One can use the DDIM~\cite{song.etal2021denoising} strategy to sample from the generative process more efficiently \begin{equation}
    \label{eq:reverse1}
    \xbm_{t-1} = \sqrt{\bar\alpha_{t-1}}\xbm_{0t} + \sqrt{1-\bar\alpha_{t-1}}(\sqrt{1-\eta_t}\epsilonbm_t + \sqrt{\eta_t}\epsilonbm),
\end{equation}
where the magnitude of $\eta_{t}=\eta\sigma_t^2/(1-\bar\alpha_{t-1})$ controlled by $\eta\in\R^+_{0}$ determines how stochastic the forward process is 
(\emph{e.g.,}  when  $\eta=0$, ~\eqref{eq:reverse1} becomes deterministic).

\medskip\noindent
\textbf{Inverse Problems.} The FVR can be formulated as an inverse problem involving the recovery of a sequence $\{\Xbm^{n}\}_{n=1}^N \in \R^{H\times W\times C}$ of video frames from a series of LQ measurements, where $N, H, W$, and $C$ are the video length, height, width, and channel, respectively. For $\xbm=[\xbm^{1}, \dots ,\xbm^{N}] \in \R^{Nd}$ defined in a vector form, we have $\xbm^{n}=\texttt{vec}(\Xbm^{n})^\Tsf\in \R^d$. The measurements can be represented as $\ybm = \Acal(\xbm) + \ebm$, where $\Acal=[\Acal_1, \dots ,\Acal_N]:\R^{Nm}\rightarrow\R^{Nd}~(m\ll d)$ is the measurement operator modeling the degradation process, and $\ebm=[\ebm^{1}, \dots,\ebm^{N}]\in\R^{Nm}$ denotes the measurement noise. In this paper, we consider the scenario in which video quality suffers from spatial and temporal degradation of images due to factors such as out-of-focus, motion, limited sensor array intensity, and JPEG encoding~\cite{Cao.etal2022towards, liang.etal2021mutual, Wang.etal2022survey}.

\section{Proposed Approach: FLAIR}
\label{sec:proposed}

In this section, we describe the training and testing details of FLAIR tailored for FVR. Fig.~\ref{Fig:fig1} illustrates the overview of the proposed method. FLAIR is defined as a generative process over $T$ steps conditioned on degraded video sequence 
 $\ybm$,
\begin{equation}
\begin{aligned}
    \label{eq:cond-reverse1}
    &\quad\quad p_{\thetabm}(\xbm_{0:T}|\ybm) = p(\xbm_T)\prod_{t=1}^{T}p_{\thetabm}(\xbm_{t-1}|\xbm_t,\ybm),  
\end{aligned}    
\end{equation}
 where $\xbm_T$ is sampled from the normal distribution $p(\xbm_T)\sim\Ncal(\zerobm,\Ibf)$, and $\xbm_0$ is the final diffusion output. Conditional generative process $p_\thetabm(\xbm_{t-1}|\xbm_t, \ybm)$ is learned to approximate the intractable conditional reverse process $q(\xbm_{t-1}|\xbm_t, \xbm_{0}, \ybm)$ for the inference, similar to unconditional DPMs. 

\subsection{Diffusion Video Restoration Network}
We leverage pre-trained DPMs for images to efficiently train the video diffusion model~\cite{Blattmann.etal2023align, Singer.etal2022make}. Our proposed method extends a DPM designed for image restoration, denoted as $\epsilon_\thetabm$, into a video diffusion restoration network represented as $\epsilon_{\thetabm,\phibm}$. 
We introduce additional~\emph{temporal}
neural network layers parameterized by $\phibm$ to $\epsilon_\thetabm$ and fine-tune them to align individual frames for temporal consistency. We adopt UNet architecture in~\cite{Dhariwal.etal2021} for network $\epsilon_\thetabm$. 
The training of the conditional model requires concatenation of the input image $\xbm_t\in\R^d$ and condition  $\cbm\in\R^d$ along the channel dimension. The condition $\cbm$ represents the up-scaled LQ measurements $\ybm\in\R^m$ to the same dimension as $\xbm_{0:T}$ (see supplements for more details). The objective function for training the $\epsilon_\thetabm$ is 
\begin{equation}
    \label{eq:condlossTheta}
    \Lcal_{\thetabm} = \E_{\xbm_0,\cbm,\epsilonbm, t\sim[1,T]}\left[\|\epsilonbm - \epsilon_{\thetabm}(\xbm_t, \cbm, t)\|^2\right].
\end{equation} 

\begin{figure}[t!]
	\centering
	\includegraphics[width=0.46\textwidth]{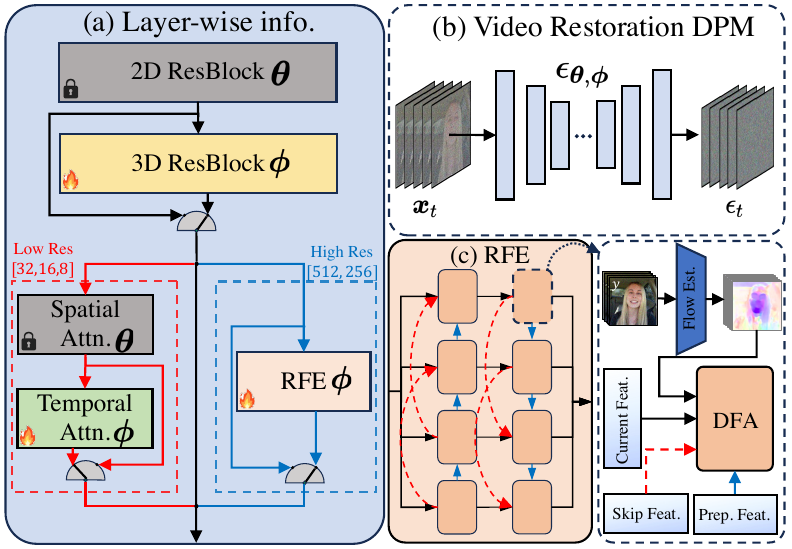}
	\caption{Overview of our video DPM.~\emph{(a)} Layer-wise information of UNet model. The image DPM backbone $\thetabm$ is fixed and only temporal layers $\bm\phi$ are fine-tuned. The recurrent feature refinement (RFE) and temporal attention are selected for different resolutions.~\emph{(c)} Illustration of the RFE module, where each recurrent block takes the flow estimation from $\ybm$ for feature alignment.}
	\label{Fig:fig2}
 \vspace{-1em}
\end{figure}

\noindent
\textbf{Temporal Layers Implementation.} Input feature maps in the pixel space are processed using the layers of image DPM denoted as~\emph{spatial} layers $\{\Hcal_{\thetabm}^i\}_{i=1}^L$, while each interleaved \emph{temporal} layer is denoted as  $\Hcal_{\bm\phi}^i$.  We use three distinct types of temporal layers depicted in Fig.~\ref{Fig:fig2}: recurrent feature enhancement (RFE), 3D convolutional residual blocks, and  temporal attention. In practice, the spatial layers $\Hcal_{\thetabm}^i$ process the video as a collection of individual images within a batch by rearranging the temporal dimension into the batch axis, i.e., $\R^{B\times C\times N \times H \times W} \rightarrow \R^{(BN)\times C \times H \times W}$, where $B$ is the batch size. Subsequently, we reshape it back to the original video dimensions for each temporal layer $\Hcal_{\bm\phi}^i$.

Directly integrating temporal attention into high-resolution features within image DPMs would notably increase memory complexity. Hence, we propose a method to capture sequential dependencies and synchronize video frame features at high resolutions (e.g., [512, 256]) by using recurrent feature refinement.
The RFE module is comprised of a 3D convolutional residual block for extracting temporal features $\tilde\fbm_i$ from the spatial output $\fbm_i$ of $\Hcal_{\thetabm}^i$ and a flow-guided deformable feature alignment (DFA) module~motivated by~\cite{Chan.etal2022basicvsr++}. The DFA is designed for bidirectional propagation, aiming to enhance the robustness of the recurrent network against error accumulation and alteration in appearance. Additional details regarding the modified DFA can be found in the supplements.

In addition, we integrate temporal attention following each $\Hcal_{\thetabm}^i$ to concurrently process $N$ frames locally in parallel  within low-resolution blocks (e.g., [32, 16, 8]). To enhance the expressiveness of modeling sequential representation, we include sinusoidal positional embeddings~\cite{Hao.etal2020} into the attention blocks. Our video temporal backbone is then trained with the same noise schedule as in~\eqref{eq:sampleNoisfwd}. We optimize the temporal layers' weights with the objective function  
\begin{equation}
    \label{eq:condlossPhi}
    \Lcal_{\phibm} = \E_{\xbm_0,\cbm,\epsilonbm, t\sim[1,T]}\left[\|\epsilonbm - \epsilon_{\thetabm, \phibm}(\xbm_t, \cbm, t)\|^2\right], 
\end{equation}  
while the spatial layers are frozen. 

\subsection{Analytical Data Consistency Module}
We initially consider a linear forward-model $\ybm^{n} = (\hbm_n*\xbm^{n})\downarrow_{s}$ without noise added to individual frames $\{\ybm^{n}\}_{n=1}^N\in\R^m$. In this expression,  $\hbm_n * \xbm^{n}$ denotes two-dimensional convolution of clean image $\xbm^{n}$ and the blur kernel associated with the point-spread function (PSF) of the camera at frame $n$ and $\downarrow_s$ represents a $s$-fold down-sampler. For convenience, we denote the forward-model as $\ybm=\Acal\xbm$. We enforce consistency of reconstructed $\tilde\xbm$ (e.g., $\xbm_{0t}$ in~\eqref{eq:x0t}) by using a projection onto the subspace spanned by $\Acal\xbm$, where $\text{rank}(\Acal)=Nm\leq Nd$. We recover the consistent reconstruction by solving the following minimization problem 
\begin{equation}
\begin{aligned}
    \label{eq:FVRFWD1}
    &\tilde\xbm_{0t} = \argmin_{\tilde\xbm}{\|\tilde\xbm - \xbm_{0t}\|_2^2}\quad\text{s.t.}\quad\Acal\tilde\xbm=\ybm,
\end{aligned}    
\end{equation}
corresponding to least-norm problem with equality constraints. This problem can be solved analytically~\cite{Boyd.Vandenberghe2004} as
\begin{equation}
\begin{aligned}
    \label{eq:rangenull}
    &\tilde\xbm_{0t} = \xbm_{0t} - \Acal^+(\Acal\xbm_{0t} - \ybm),
\end{aligned}    
\end{equation}
where $\Acal^+=\Acal^{T}(\Acal\Acal^{T})^{-1}\in\R^{Nd\times Nm}$ is the Moore-Penrose pseudo-inverse of $\Acal$ and satisfies $\Acal\Acal^+=\Ibf_{Nm}$.
By substituting the estimated $\xbm_{0t}$ with $\tilde\xbm_{0t}$ in~\eqref{eq:reverse1}, we enforce the low-frequency content of $\tilde\xbm_{0t}$ to align with that of the ground-truth video sequence $\xbm$ (\emph{i.e.,} $\Acal\tilde\xbm_{0t}=\Acal\xbm=\ybm$), while allowing the reverse diffusion process to recover the high-frequency components.
We reformulate ~\eqref{eq:rangenull} by calculating $\Acal^+$ according to~\cite{bahat2020explorable} for each individual frame 
\begin{equation*}
\begin{aligned}
    \label{eq:srclosed}
    &\tilde\xbm_{0t}^{n} = \xbm_{0t}^{n} - \tilde\hbm_n * \left(\kbm_n*((\hbm_n*\xbm_{0t}^{n})\downarrow_{s}-\ybm^{n})\right)\uparrow^s,\\
\end{aligned}    
\end{equation*}
where $\tilde\hbm_n$ is the mirrored version of the blur kernel $\hbm_n$, and $\uparrow^s$ denotes spatial upsampling by zero-filling of new entries.
$\kbm_n$ is used to replace the multiplication by $(\Acal\Acal^{T})^{-1}$ and corresponds to the inverse of filter $(\hbm_n*\tilde\hbm_n)\downarrow_s$ in Fourier domain. 

\begin{table*}[t!]
\centering
\resizebox{0.89\textwidth}{!}{%
\begin{tabular}{lccccccc|cccccc}
\hline
\multirow{2}{*}{Method} & \multirow{2}{*}{Task}                          & \multicolumn{6}{c|}{CelebV-Text~\cite{Yu2023.etalcelebv}}                      & \multicolumn{6}{c}{CelebV-HQ~\cite{Zhu.etal2022celebv}}                            \\ \cline{3-14} 
                        &                                                & PSNR$\uparrow$  & SSIM$\uparrow$   & LPIPS$\downarrow$  & FVD$\downarrow$    & FID$\downarrow$   & KID$\downarrow$   & PSNR$\uparrow$  & SSIM$\uparrow$   & LPIPS$\downarrow$  & FVD$\downarrow$     & FID$\downarrow$    & KID$\downarrow$    \\ \hline\hline

$\Acal^+ \ybm$                    & \multirow{8}{*}{\rotatebox[origin=c]{90}{$8\times$ Bicubic}}                    & 21.40  & 0.740 & 0.412 & 481.22 & 202.14 & 218.43 & 22.25 & 0.731 & 0.424 & 863.97  & 256.04 & 257.19  \\
VQFR~\cite{Gu.etal2022vqfr}                    &                    & 26.40  & 0.801 & 0.255 & 229.86 & 76.46 & 23.24 & 25.81 & 0.777 & 0.277 & 482.91  & 126.86 & 41.15  \\
RestoreFormer++~\cite{Wang2023.etalrestoreformer++}           &                                                & 26.48 & 0.799 & 0.249 & 190.48 & 70.39 & 16.81 & 25.98 & 0.775 & 0.273 & 470.12  & 123.14 & 38.55  \\
CodeFormer~\cite{Zhou.etal2022towards}               &                                                & 26.66 & 0.798 & 0.259  & 214.37 & 76.11 & 21.39 & 26.00    & 0.775 & 0.278 & 498.19  & 126.28 & 39.34  \\
DR2E~\cite{Wang.etal2023dr2}                     &                                                & 27.89 & 0.824 & \colorbox{lightblue}{\makebox(24,4){0.202}} & 205.48 & 53.68 & \colorbox{lightblue}{\makebox(24,4){13.51}} & 27.49 & 0.8073 & \colorbox{lightblue}{\makebox(24,4){0.207 }} & 419.64  & 91.28  & \colorbox{lightblue}{\makebox(24,4){22.86}}  \\
DDNM~\cite{Wang.etal2023zeroshot}                    &                                                & \colorbox{lightblue}{\makebox(24,4){29.95}} & \colorbox{lightblue}{\makebox(24,4){0.860}} & 0.234 & \colorbox{lightblue}{\makebox(24,4){122.03}} & 72.16 & 44.07 & \colorbox{lightblue}{\makebox(24,4){29.00}}    & \colorbox{lightblue}{\makebox(24,4){0.836}} & 0.253 & 352.08  & 113.65 & 64.10   \\
ILVR~\cite{Choi.etal2021}                    &                                                & 29.62 & 0.852 & 0.206 & 145.22 & \colorbox{lightblue}{\makebox(24,4){52.72}} & 21.39 & 28.77 & 0.829 & 0.222 & \colorbox{lightblue}{\makebox(24,4){350.38}}  & \colorbox{lightblue}{\makebox(24,4){90.95}}  & 37.88  \\
FLAIR (Ours)                     &                                                & \colorbox{lightgreen}{\makebox(24,4){\bf30.76}} & \colorbox{lightgreen}{\makebox(24,4){\bf0.868}} & \colorbox{lightgreen}{\makebox(24,4){\bf0.159}} & \colorbox{lightgreen}{\makebox(24,4){\bf75.16}}  & \colorbox{lightgreen}{\makebox(24,4){\bf41.46}} & \colorbox{lightgreen}{\makebox(24,4){\bf8.11}}  & \colorbox{lightgreen}{\makebox(24,4){\bf29.56}} & \colorbox{lightgreen}{\makebox(24,4){\bf0.844}} & \colorbox{lightgreen}{\makebox(24,4){\bf0.157}} & \colorbox{lightgreen}{\makebox(24,4){\bf194.79}}  & \colorbox{lightgreen}{\makebox(24,4){\bf66.69}} & \colorbox{lightgreen}{\makebox(24,4){\bf13.79}}  \\ \hline
$\Acal^+ \ybm$                    & \multirow{8}{*}{\rotatebox[origin=c]{90}{$16\times$ Bicubic}}                   & 20.81 & 0.721 & 0.542 & 1278.46 & 182.22 & 150.72 & 21.32 & 0.704 & 0.567 & 2145.50  & 260.01 & 183.21   \\
VQFR~\cite{Gu.etal2022vqfr}                    &                    & 23.49 & 0.746 & 0.362 & 500.76 & 97.27 & 32.12 & 22.78 & 0.716 & 0.407 & 1103.10  & 180.93 & 59.50   \\
RestoreFormer++~\cite{Wang2023.etalrestoreformer++}           &                                                & 23.29 & 0.732 & 0.368 & 518.95 & 92.86 & 26.20  & 22.75 & 0.706 & 0.414 & 1154.06 & 175.27 & 50.04  \\
CodeFormer~\cite{Zhou.etal2022towards}               &                                                & 23.58 & 0.738 & 0.374 & 507.03 & 101.20 & 32.90  & 22.89 & 0.711 & 0.419 & 1155.91 & 178.08 & 55.84  \\
DR2E~\cite{Wang.etal2023dr2}                     &                                                & 24.38 & 0.755 & 0.314 & \colorbox{lightblue}{\makebox(24,4){456.12}} & 81.42 & \colorbox{lightblue}{\makebox(24,4){21.81}} & 23.73 & 0.726 & 0.349 & \colorbox{lightblue}{\makebox(24,4){984.00}}     & 148.80  & \colorbox{lightblue}{\makebox(24,4){37.55}} \\
DDNM~\cite{Wang.etal2023zeroshot}                    &                                                & \colorbox{lightblue}{\makebox(24,4){25.78}} & \colorbox{lightblue}{\makebox(24,4){0.789}} & 0.337 & 617.05 & 82.07 & 41.80  & \colorbox{lightblue}{\makebox(24,4){24.85}} & \colorbox{lightblue}{\makebox(24,4){0.753}}& 0.368 & 1264.72 & 148.13 & 67.27  \\
ILVR~\cite{Choi.etal2021}                    &                                                & 25.56 & 0.777 & \colorbox{lightblue}{\makebox(24,4){0.285}} & 635.46 & \colorbox{lightblue}{\makebox(24,4){68.90}}  & 23.83 & 24.74 & 0.743 & \colorbox{lightblue}{\makebox(24,4){0.312}}& 1306.38 & \colorbox{lightblue}{\makebox(24,4){128.67}}& 47.93  \\
FLAIR (Ours)                     &                                                & \colorbox{lightgreen}{\makebox(24,4){\bf26.70}}  & \colorbox{lightgreen}{\makebox(24,4){\bf0.800}}    & \colorbox{lightgreen}{\makebox(24,4){\bf0.216}} & \colorbox{lightgreen}{\makebox(24,4){\bf158.05}} & \colorbox{lightgreen}{\makebox(24,4){\bf58.09}} & \colorbox{lightgreen}{\makebox(24,4){\bf8.94}}  & \colorbox{lightgreen}{\makebox(24,4){\bf25.49}} & \colorbox{lightgreen}{\makebox(24,4){\bf0.758}} & \colorbox{lightgreen}{\makebox(24,4){\bf0.222}} & \colorbox{lightgreen}{\makebox(24,4){\bf442.55}}  & \colorbox{lightgreen}{\makebox(24,4){\bf99.15}}  & \colorbox{lightgreen}{\makebox(24,4){\bf17.56}}  \\ \hline
$\Acal^+ \ybm$                    & \multirow{8}{*}{\rotatebox[origin=c]{90}{\makecell{$4\times$, Gaussian blur \\ $\sigma=0.05$ }}} & 17.21 & 0.287 & 0.832 & 1905.12 & 143.77 & 85.97  & 17.77 & 0.299 & 0.827 & 3022.43  & 204.81  & 108.90  \\
VQFR~\cite{Gu.etal2022vqfr}                    & 
& 27.54 & 0.810 & 0.195 & 385.35 & 50.22 & 9.62  & 27.87 & 0.816 & 0.200 & 628.88  & 84.72  & 16.94  \\
RestoreFormer++~\cite{Wang2023.etalrestoreformer++}           &                                                & 28.13 & 0.818 & \colorbox{lightblue}{\makebox(24,4){0.193}}  & 322.94 & 47.15 & \colorbox{lightblue}{\makebox(24,4){7.99}}& 27.90  & 0.813 & \colorbox{lightblue}{\makebox(24,4){0.191}} & 527.08  & \colorbox{lightblue}{\makebox(24,4){78.29}}  & \colorbox{lightgreen}{\makebox(24,4){\bf13.85}} \\
CodeFormer~\cite{Zhou.etal2022towards}               &                                                & 28.64 & 0.825 & 0.193 & \colorbox{lightblue}{\makebox(24,4){294.92}} & 50.09 & 9.09  & 28.04  & 0.816 & 0.192 & \colorbox{lightblue}{\makebox(24,4){494.30}}   & 81.99  & 15.65  \\
DR2E~\cite{Wang.etal2023dr2}                     &                                                & 27.43 & 0.802 & 0.220   & 564.43 & 56.15 & 12.45 & 27.01 & 0.788 & 0.218 & 909.62  & 100.89 & 20.86  \\
DDNM~\cite{Wang.etal2023zeroshot}                    &                                                & \colorbox{lightgreen}{\makebox(24,4){\bf30.24}} & \colorbox{lightgreen}{\makebox(24,4){\bf0.863}} & 0.250   & 320.77 & 74.11 & 34.40  & \colorbox{lightgreen}{\makebox(24,4){\bf29.20}}   & \colorbox{lightgreen}{\makebox(24,4){\bf0.846}} & 0.265 & 629.74  & 112.86 & 50.14  \\
DiffPIR~\cite{Zhu.etal2023denoising}                 &                                                & 28.93 & 0.838 & 0.210 & 672.55 & \colorbox{lightblue}{\makebox(24,4){\colorbox{lightblue}{\makebox(24,4){43.80}}}}  & \colorbox{lightgreen}{\makebox(24,4){\bf6.29}}  & 28.04 & 0.815  & 0.223 & 1051.06 & 83.07  & 16.86  \\
FLAIR (Ours)                     &                                                & \colorbox{lightblue}{\makebox(24,4){29.87}} & \colorbox{lightblue}{\makebox(24,4){0.856}}  & \colorbox{lightgreen}{\makebox(24,4){\bf0.149}} & \colorbox{lightgreen}{\makebox(24,4){\bf82.82}} & \colorbox{lightgreen}{\makebox(24,4){\bf39.54}} & 8.25 & \colorbox{lightblue}{\makebox(24,4){28.15}} & \colorbox{lightblue}{\makebox(24,4){0.818}}& \colorbox{lightgreen}{\makebox(24,4){\bf0.179}}& \colorbox{lightgreen}{\makebox(24,4){\bf255.44}}

 & \colorbox{lightgreen}{\makebox(24,4){\bf74.47}}& \colorbox{lightblue}{\makebox(24,4){14.40}}   \\ \hline
 $\Acal^+ \ybm$                    & \multirow{7}{*}{\rotatebox[origin=c]{90}{\makecell{$4\times$, Gaussian blur\\$\sigma=0.05$, JPEG$60$}}} & 19.53 & 0.481 & 0.710 & 1856.50 & 141.39 & 90.56  & 20.15 & 0.472 & 0.696 & 2990.26  & 205.48  & 113.83  \\
VQFR~\cite{Gu.etal2022vqfr}                    & 
& 27.15 & 0.807 & 0.214 & 483.55 & 54.09 & 10.59  & 26.68 & 0.798 & 0.215 & 807.43  & 94.40  & 19.59  \\
RestoreFormer++~\cite{Wang2023.etalrestoreformer++}           &                                                & 27.12 & 0.806 & 0.214  & 427.63 & \colorbox{lightblue}{\makebox(24,4){52.58}} & \colorbox{lightblue}{\makebox(24,4){9.42}} & 26.83  & 0.797 & \colorbox{lightblue}{\makebox(24,4){0.214}}& 739.01  & \colorbox{lightblue}{\makebox(24,4){89.94}} & \colorbox{lightblue}{\makebox(24,4){17.20}}\\
CodeFormer~\cite{Zhou.etal2022towards}               &                                                & 27.71 & 0.814 & \colorbox{lightblue}{\makebox(24,4){\colorbox{lightblue}{\makebox(24,4){0.211}}}} & \colorbox{lightblue}{\makebox(24,4){385.63}} & 55.24 & 10.74  & 27.05  & 0.802 & 0.215 & \colorbox{lightblue}{\makebox(24,4){720.54}} & 94.25  & 19.15  \\
DR2E~\cite{Wang.etal2023dr2}                     &                                                & 26.58 & 0.789 & 0.242   & 695.99 & 60.39 & 12.89 & 26.01 & 0.773 & 0.243 & 1091.43  & 116.38 & 21.90  \\
DDNM~\cite{Wang.etal2023zeroshot}                    &                                                & \colorbox{lightblue}{\makebox(24,4){\colorbox{lightblue}{\makebox(24,4){29.02}}}} & \colorbox{lightblue}{\makebox(24,4){\colorbox{lightblue}{\makebox(24,4){0.851}}}} & 0.271   & 509.15 & 74.48 & 35.89  & \colorbox{lightblue}{\makebox(24,4){27.63}} & \colorbox{lightblue}{\makebox(24,4){0.818}} & 0.317 & 1067.57  & 126.23 & 57.91  \\
FLAIR (Ours)                     &                                                & \colorbox{lightgreen}{\makebox(24,4){\bf29.39}} & \colorbox{lightgreen}{\makebox(24,4){\bf0.857}}  & \colorbox{lightgreen}{\makebox(24,4){\bf0.178}} & \colorbox{lightgreen}{\makebox(24,4){\bf126.36}}& \colorbox{lightgreen}{\makebox(24,4){\bf45.90}} & \colorbox{lightgreen}{\makebox(24,4){\bf9.11}} & \colorbox{lightgreen}{\makebox(24,4){\bf28.40}} & \colorbox{lightgreen}{\makebox(24,4){\bf0.841}} & \colorbox{lightgreen}{\makebox(24,4){\bf0.185}} & \colorbox{lightgreen}{\makebox(24,4){\bf316.89}} & \colorbox{lightgreen}{\makebox(24,4){\bf74.12}} & \colorbox{lightgreen}{\makebox(24,4){\bf14.04}} \\ \hline
\end{tabular}%
}
\vspace{-0.5em}
\caption{
Quantitative results on two face video datasets (short clips). \colorbox{lightgreen}{\makebox(20,5.5){\textbf{Best}}}~and~\colorbox{lightblue}{\makebox(40,5.5){second-best}} values for each metric are color-coded.}
\label{tab:table1}
\vspace{-0.5em}
\end{table*}
\begin{figure*}[t!]
	\centering
	\includegraphics[width=0.95\textwidth]{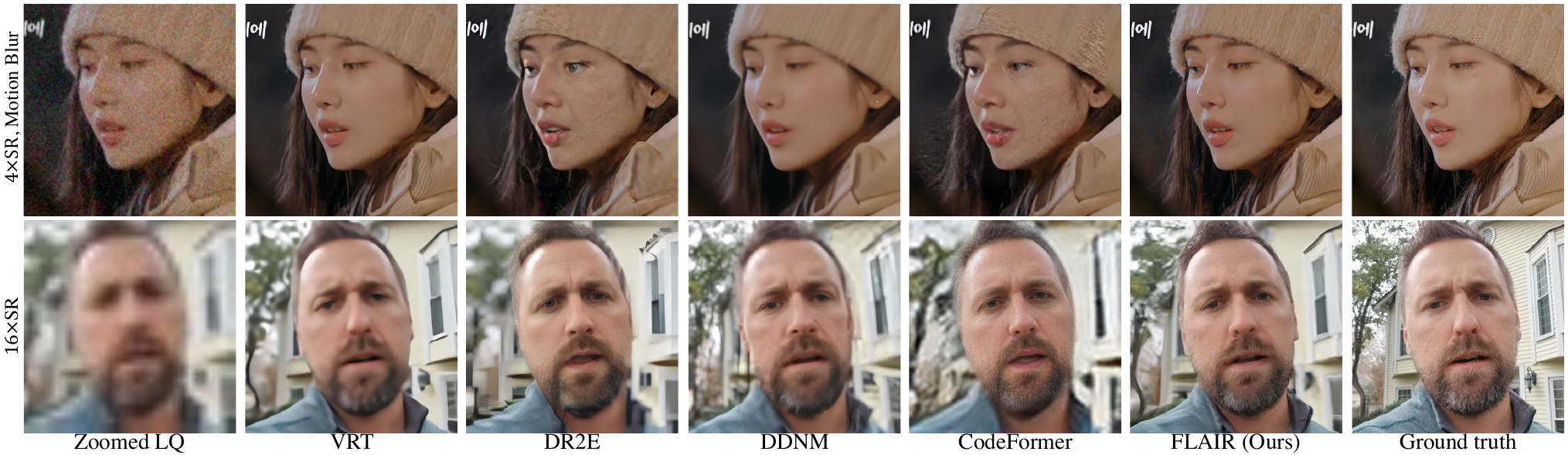}
    \vspace{-1em}
	\caption{Visual comparisons. Our FLAIR produces higher restoration quality while
maintaining data fidelity well.}
	\label{Fig:fig4}
 \vspace{-.5em}
\end{figure*}
\begin{figure*}[t!]
	\centering
	\includegraphics[width=0.95\textwidth]{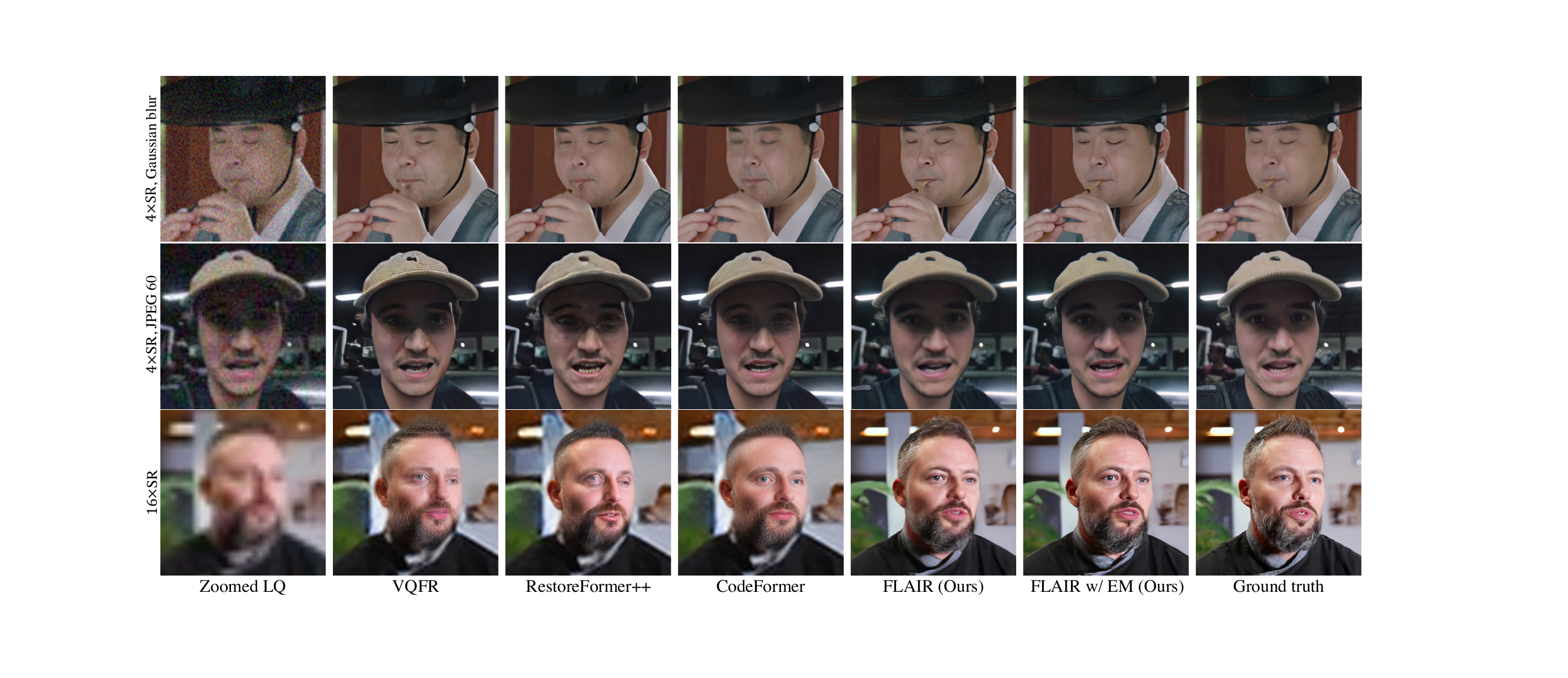}
    \vspace{-1em}
	\caption{Qualitative comparisons. Our method with different enhancement module backbones achieve higher restoration quality while maintaining data fidelity well. ~\emph{\textbf{Top}}: FLAIR + RestoreFormer++.~\emph{\textbf{Middle}} and ~\emph{\textbf{Bottom}}: FLAIR + CodeFormer.}
	\label{Fig:fig5}
 \vspace{-0.5em}
\end{figure*}

\noindent
\textbf{Noisy FVR Degradation.}
In the presence of noise, the forward model is $\ybm=\Acal\xbm +\ebm$, where $\ebm=\{\ebm^{n}\}_{n=1}^N$ represents additive white Gaussian noise (AWGN) with $\ebm^{n}\sim \Ncal(\zerobm,\sigma^2_\ebm\Ibf_m)$. Directly applying ~\eqref{eq:rangenull} to noisy measurements $\ybm$ will result in an additional noise term $\Acal^+\ebm$ in $\tilde\xbm_{0t}$, consequently affecting the reverse diffusion $q(\xbm_{t-1}|\xbm_t, \tilde\xbm_{0}, \ybm)$. We can approximate $\Acal^+_n\ebm^{n}$ as a AWGN $\Ncal(\zerobm, \sigma^2_\ebm\Ibf_m)$, given $\Acal^+$ in FVR closely resembles a copy operation~\cite{Wang.etal2023zeroshot}. Thus~\eqref{eq:rangenull} and~\eqref{eq:reverse1} can be modified into
\begin{align}
    \label{eq:rangenullscale}
    &\tilde\xbm_{0t} = \xbm_{0t} - \gamma_t\Acal^+(\Acal\xbm_{0t} - \Acal\xbm) + \gamma_t\Acal^+\ebm,\\\nonumber    
    &\xbm_{t-1}=\sqrt{\bar\alpha_{t-1}}\tilde\xbm_{0t} + \sqrt{1-\bar\alpha_t}(\sqrt{1-\rho_t}\tilde\epsilonbm_t + \sqrt{\rho_t}\epsilonbm),  
\end{align}  
where $\gamma_t\geq0$ and $\rho_t>0$ are user-defined hyperparameters such that $\sigma_t=\sqrt{\bar\alpha_{t-1}\gamma_t^2\sigma_\ebm^2 + \rho_t}$, and $\tilde\epsilonbm_t=\frac{1}{\sqrt{1-\bar\alpha_t}}(\xbm_t - \sqrt{\bar\alpha_t}\tilde\xbm_{0t})\in\R^{Nd}$ is the recalculated noise estimate. 
By appropriately setting $\gamma_t$ and $\rho_t$, we make the total noise variance in $\xbm_{t-1}$ conform to the forward diffusion $q (\xbm_{t-1}|\xbm_0)$ in~\eqref{eq:sampleNoisfwd}. This allows for an effective estimation of noise by $\epsilon_{\thetabm,\phibm}(\xbm_{t-1},\cbm,t)$ at next step.

\noindent
\textbf{Composite FVR Degradation.}
FLAIR is also applicable to more complicated FVR degradation
\begin{equation}
\begin{aligned}
    \label{eq:complexFVR}
    &\ybm^{n} = \Ecal_n\left((\hbm_n*\xbm^{n})\downarrow_{s} + \ebm^n\right),
\end{aligned}
\end{equation}
where $\Ecal=\{\Ecal_n\}_{n=1}^{N}$ denotes the JPEG encoding with quality factors $Q\geq0$. While JPEG is non-linear, we can construct JPEG decoding operator $\Dcal$, such that $\Ecal(\Dcal(\Ecal(\xbm)))=\Ecal(\xbm), \forall\xbm\in\R^{Nn}$, similar to~\cite{Kawar.etal2022jpeg}, which is analogue to the matrix pseudo-inverse $\Acal\Acal^{+}\Acal\xbm=\Acal\xbm$. 
For composite forward operator $\Acal=\Acal_1\circ...\circ\Acal_k$, we may approximate $\Acal^{+}$ with $\Acal^{+}=\Acal_k^{+}\circ...\circ\Acal_1^{+}$. Hence, 
$\tilde\xbm_{0t}$ in~\eqref{eq:rangenullscale} under composite degradation in~\eqref{eq:complexFVR} can be efficiently solved using
\begin{align}
    \label{eq:compositrangenullscale}
    &\tilde\xbm_{0t} = \xbm_{0t} - \gamma_t\Acal^+\Dcal(\Ecal(\Acal\xbm_{0t}) - \ybm).
\end{align}
The full algorithm of FLAIR is detailed in the supplements.

\subsection{Efficient Spatial Enhancement Module}
Finally, we  introduce a coarse-to-fine image enhancement module designed for refinement of estimated $\tilde\xbm_{0t}$, as
\begin{equation}
\begin{aligned}
    \label{eq:enhance}
    &\tilde\xbm_{0t}= (\bm{1}-w_t \bm m_t)\odot\tilde\xbm_{0t} + w_t \bm m_t\odot\mathcal{G}(\tilde\xbm_{0t}),
\end{aligned}
\end{equation}
where $w_t$ balances the importance of the facial enhancement region $\bm m\odot\mathcal{G}(\tilde\xbm_{0t})\in\R^{Nd}$ and originally estimated $\bm{m}_t\odot\tilde\xbm_{0t}\in\R^{Nd}$ at each step, and $(\bm{1}-\bm{m}_t)\odot\tilde\xbm_{0t}$ denotes the background scenes. Note that we~\emph{do not} impose any specific constraints on the method or architecture of $\mathcal{G}$, allowing the enhancement module to be trained independently.
For our enhancement module, we consider two well-established backbones: Restorformer++~\cite{Wang2023.etalrestoreformer++} and Codeformer~\cite{Zhou.etal2022towards}. This shows the compatibility of FLAIR with a diverse range of existing methods. Both backbones make use of pre-trained high-quality VQ codebooks~\cite{Esser.etal2021taming} specifically designed for face images. We refer to these methods as \emph{FLAIR + RestorFormer++} and \emph{FLAIR + CodeFormer}, respectively.

\section{Experiments}
\label{Sec:Experiments}
\subsection{Experimental Setup}
\label{Sec:Experimental}
\noindent
\textbf{Datasets.}
We use FFHQ~\cite{karras.etal2019style} for training image DPMs and 7200 clips from CelebV-Text~\cite{Yu2023.etalcelebv} for fine-tuning video DPMs. We choose 125 short clips and 6 long clips from the unused identities of the CelebV-Text for testing. We also consider 20 clips from CelebV-HQ~\cite{Zhu.etal2022celebv} and 100 sequences from Obama datasets~\cite{Suwajanakorn.etal2017synthesizing} for testing. We additionally crawl a real life video clip with 300 frames from the Internet for testing. See supplements for more details.

\noindent
\textbf{Evaluation Metrics.} Our evaluation is based on both perception and distortion of the restored videos. For perception, we choose three different frame-wise perceptual metrics: Frechet Inception Distance (FID)~\cite{Heusel.etal2017gans}, LPIPS~\cite{Zhang.etal2018unreasonable}, and Kernel Inception Distance (KID)~\cite{Binkowski.etal2018demystifying} as well as Frechet Video Distance (FVD)~\cite{Unterthiner.etal2018towards}. We adopt two pixel-wise metrics: PSNR and SSIM~\cite{Wang.etal2004} to evaluate data fidelity of our method.

\noindent
\textbf{Training and Inference Details.}
We consider three types of degradation models: video super-resolution (SR), deblurring and JPEG restoration. For video SR, we pre-train a conditional image DPM backbone (spatial layers) using downsampling factors $s=8$ with bicubic degradation and then fine tune the video DPMs with loss function in~\eqref{eq:condlossPhi} using $s\in\{4, 8, 16\}$ separately. Likewise, for video deblurring, we pre-train a conditional image DPM for our video DPM using scale factors $s=4$ and AWGN $\sigma_\ebm\in[0,25]$ with anisotropic Gaussian kernels as in~\cite{Riegler2015.etalconditioned, zhang.etal2020a} and motion kernels as in~\cite{Boracchi.etal2012modeling}. We fix the kernel size to $25\times25$. For video JPEG restoration, we use the same settings as for deblurring with additional JPEG quality factor $Q\in[60, 100]$. 

\begin{table}[]
\centering
\resizebox{1.\columnwidth}{!}{%
\begin{tabular}{lcccccc}
\hline
Method             & PSNR$\uparrow$  & SSIM$\uparrow$ & LPIPS$\downarrow$  & FVD$\downarrow$     & FID$\downarrow$    & KID$\downarrow$   \\ \hline
\multicolumn{7}{c}{$16\times$ Bicubic}                                                           \\ \hline
$\Acal^+ \ybm$                & 16.89 & 0.657                   & 0.621 & 4456.46  & 216.34 & 145.39 \\
VRT~\cite{Liang.etal2022vrt}                & \colorbox{lightgreen}{\makebox(24,4){\bf29.38}} & \colorbox{lightgreen}{\makebox(24,4){\bf0.866}}                   & 0.287 & 580.65  & 114.91 & 54.26 \\
BasicVSRPP~\cite{Chan.etal2022basicvsr++}         & 26.91 & 0.836                   & 0.308 & 634.21  & 143.10  & 63.24 \\
CodeFormer~\cite{Zhou.etal2022towards}          & 23.80  & 0.738                   & 0.366  & 1059.39 & 141.79 & 51.20  \\
RestoreFormer++~\cite{Wang2023.etalrestoreformer++}      & 23.80  & 0.742                   & 0.353 & 518.95  & 92.86  & 26.20  \\
VQFR~\cite{Gu.etal2022vqfr}               & 23.62 & 0.739                   & 0.356  & 1019.93 & 141.12 & 48.92 \\
DR2E~\cite{Wang.etal2023dr2}                & 24.83 & 0.764                   & 0.312 & 903.16  & 113.63 & 35.93 \\
DDNM~\cite{Wang.etal2023zeroshot}               & 26.28 & 0.809                   & 0.343 & 846.88  & 104.91 & 48.88 \\ \hdashline
FLAIR (Ours)               & 28.23 & 0.842                   & 0.240 & 358.72  & 84.40   & 27.26 \\
FLAIR-SA (Ours)   & \colorbox{lightblue}{\makebox(24,4){28.96}} & \colorbox{lightblue}{\makebox(24,4){0.855}}                   & 0.268 & 571.36  & 95.90   & 35.97 \\
FLAIR+CodeFormer (Ours)    & 27.57 & 0.830                   & \colorbox{lightgreen}{\makebox(24,4){\bf0.212}} & \colorbox{lightgreen}{\makebox(24,4){\bf344.99}}  & \colorbox{lightblue}{\makebox(24,4){80.47}}  & \colorbox{lightblue}{\makebox(24,4){24.71}} \\
FLAIR+RestoreFormer++ (Ours) & 27.31 & 0.819                    & \colorbox{lightblue}{\makebox(24,4){0.233}} & \colorbox{lightblue}{\makebox(24,4){352.00}}     & \colorbox{lightgreen}{\makebox(24,4){\bf78.68}}  & \colorbox{lightgreen}{\makebox(24,4){\bf23.78}} \\ \hline
\multicolumn{7}{c}{$4\times$, Motion blur, $\sigma=0.05$}                                                        \\ \hline
$\Acal^+ \ybm$                & 14.62 & 0.244                   & 0.850 & 3515.79  & 200.59  & 134.43 \\
VRT~\cite{Liang.etal2022vrt}                & 30.58 & \colorbox{lightgreen}{\makebox(24,4){\bf0.904}}                   & 0.173 & 149.73  & 68.94  & 26.95 \\
CodeFormer~\cite{Zhou.etal2022towards}          & 27.74 & 0.817                   & 0.188 & 596.37  & 65.90   & 19.70  \\
RestoreFormer++~\cite{Wang2023.etalrestoreformer++}      & 27.88 & 0.819                    & 0.189 & 587.97  & 64.66  & 18.25 \\
VQFR~\cite{Gu.etal2022vqfr}               & 27.21 & 0.808                   & 0.205 & 836.61  & 75.00     & 22.84 \\
DR2E~\cite{Wang.etal2023dr2}                & 27.04 & 0.799                   & 0.213 & 1135.91 & 76.98  & 22.72 \\
DiffPIR~\cite{Zhu.etal2023denoising}                & 29.55 & 0.855                   & 0.213 & 1139.93 & 51.59  & 12.22 \\
DDNM~\cite{Wang.etal2023zeroshot}               & 29.21 & 0.847                   & 0.267 & 762.26  & 95.58  & 42.09 \\ \hdashline
FLAIR (Ours)               & 31.10  & 0.890                     & 0.151 & \colorbox{lightgreen}{\makebox(24,4){\bf126.24}}  & 48.21  & 15.36 \\
FLAIR-SA (Ours) & \colorbox{lightgreen}{\makebox(24,4){\bf31.66}} & \colorbox{lightblue}{\makebox(24,4){0.897}}                   & 0.152 & 131.54  & 49.68  & 17.97 \\
FLAIR+CodeFormer (Ours)    & \colorbox{lightblue}{\makebox(24,4){31.12}} & 0.891                   & \colorbox{lightblue}{\makebox(24,4){0.147}} & \colorbox{lightblue}{\makebox(24,4){127.43}}  & \colorbox{lightblue}{\makebox(24,4){47.17}}  & \colorbox{lightblue}{\makebox(24,4){14.89}} \\
FLAIR+RestoreFormer++ (Ours) & 31.03 & 0.876                   & \colorbox{lightgreen}{\makebox(24,4){\bf0.146}} & 134.59  & \colorbox{lightgreen}{\makebox(24,4){\bf43.66}}  & \colorbox{lightgreen}{\makebox(24,4){\bf12.25}} \\ \hline
\multicolumn{7}{c}{$4\times$, Gaussian blur, $\sigma=0.05$, JPEG $60$}                                                           \\ \hline
$\Acal^+ \ybm$                & 16.11 & 0.426                   & 0.728 & 3574.66  & 189.92 & 126.43 \\
CodeFormer~\cite{Zhou.etal2022towards}         & 28.58  & 0.824                   & 0.203  & 698.86 & 73.96 & 22.17  \\
RestoreFormer++~\cite{Wang2023.etalrestoreformer++}      & 28.15  & 0.818                   & 0.213 & 761.40  & 78.68  & 22.57  \\
VQFR~\cite{Gu.etal2022vqfr}               & 27.63 & 0.812                   & 0.213  & 888.53 & 76.96 & 23.20 \\
DR2E~\cite{Wang.etal2023dr2}                & 24.83 & 0.764                   & 0.312 & 903.16  & 113.63 & 35.93 \\
DDNM~\cite{Wang.etal2023zeroshot}               & 29.72 & 0.849                   & 0.275 & 954.21  & 99.86 & 50.78 \\\hdashline
FLAIR (Ours)               & \colorbox{lightblue}{\makebox(24,4){29.99}} & \colorbox{lightblue}{\makebox(24,4){0.860}}                   & 0.175 & \colorbox{lightblue}{\makebox(24,4){235.86}}  & \colorbox{lightblue}{\makebox(24,4){62.10}}  & 17.73 \\
FLAIR-SA (Ours)          & \colorbox{lightgreen}{\makebox(24,4){\bf30.57}}  &  \colorbox{lightgreen}{\makebox(24,4){\bf0.873}} & 0.199 & 295.60 & 77.17 & 30.51 \\
FLAIR+CodeFormer (Ours)     & 29.96 & 0.858                   & \colorbox{lightblue}{\makebox(24,4){0.174}} & 246.77  & \colorbox{lightgreen}{\makebox(24,4){\bf59.48}}  & \colorbox{lightgreen}{\makebox(24,4){\bf16.49}}\\
FLAIR+RestoreFormer++ (Ours) & 29.95 & 0.857                    & \colorbox{lightgreen}{\makebox(24,4){\bf0.172}} & \colorbox{lightgreen}{\makebox(24,4){\bf229.69}} & 62.27  & \colorbox{lightblue}{\makebox(24,4){17.48}} \\ \hline
\end{tabular}%
}
\caption{Quantitative results on CelebV-Text~\cite{Yu2023.etalcelebv} (long clips). \colorbox{lightgreen}{\makebox(20,5.5){\textbf{Best}}}~and~\colorbox{lightblue}{\makebox(40,5.5){second-best}} values for each metric are color-coded.}
\label{tab:table2}
\vspace{-2em}
\end{table}
We use pre-trained SPyNet~\cite{Ranjan.etal2017optical}
as our flow estimation network. At inference, we use $K\in[1,T)$ evenly spaced real numbers for the sampling step index, and then round each resulting number to the nearest integer following~\cite{Dhariwal.etal2021}. For the enhancement module, we employ the original pre-trained models of RestoreFormer~\cite{Wang2023.etalrestoreformer++} and CodeFormer~\cite{Zhou.etal2022towards}.

\subsection{Comparisons with SOTA Methods}
\label{Sec:comparisons}
We present quantitative comparisons between our FLAIR and several methods across various degradation settings in Table~\ref{tab:table1} and Table~\ref{tab:table2}. VQFR~\cite{Gu.etal2022vqfr}, CodeFormer~\cite{Zhou.etal2022towards} and RestoreFormer++~\cite{Wang2023.etalrestoreformer++} are three SOTA face restoration methods that use pre-trained high-quality facial dictionary priors. Their official released models are adopted in the experiments. Since, to the best of our knowledge, there is no existing work that uses video diffusion models for FVR, we compare FLAIR with some of the latest conditional image DPMs that use unconditionally trained diffusion models for solving inverse problems, including ILVR~\cite{Choi.etal2021}, DR2E~\cite{Wang.etal2023dr2}, DDNM~\cite{Wang.etal2023zeroshot} and DiffPIR~\cite{Zhu.etal2023denoising}. DR2E consists of a degradation removal module built upon image DPM and an enhancement module similar to FLAIR. Following the original setup, we use VQFR as the enhancement module for DR2E. For the DPM baselines, we pre-train an unconditional image DPM on FFHQ and then fine tune it on the same CelebV-Text images used for training FLAIR. For each task, we omit any method that was not implemented in the original work for fair comparison. 
\begin{figure}[t!]
	\centering
	\includegraphics[width=0.45\textwidth]{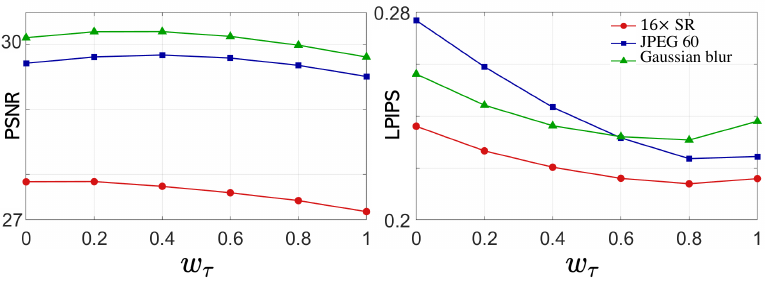}
    \vspace{-0.5em}
	\caption{Comparison of average PSNR $\uparrow$ (left) and LPIPS $\downarrow$
(right) of FLAIR with various controlling schedules $\{w_t\}_{t=\tau}^{K-1}$ in~\eqref{eq:enhance}, where $K=25$ and $\tau=5$ for all experiments. Note the improved perception (LPIPS) quality by increasing $w_{\tau}$.}
	\label{Fig:fig6}
\vspace{-1.em}
\end{figure}
\begin{figure*}[t!]
	\centering
	\includegraphics[width=0.98\textwidth]{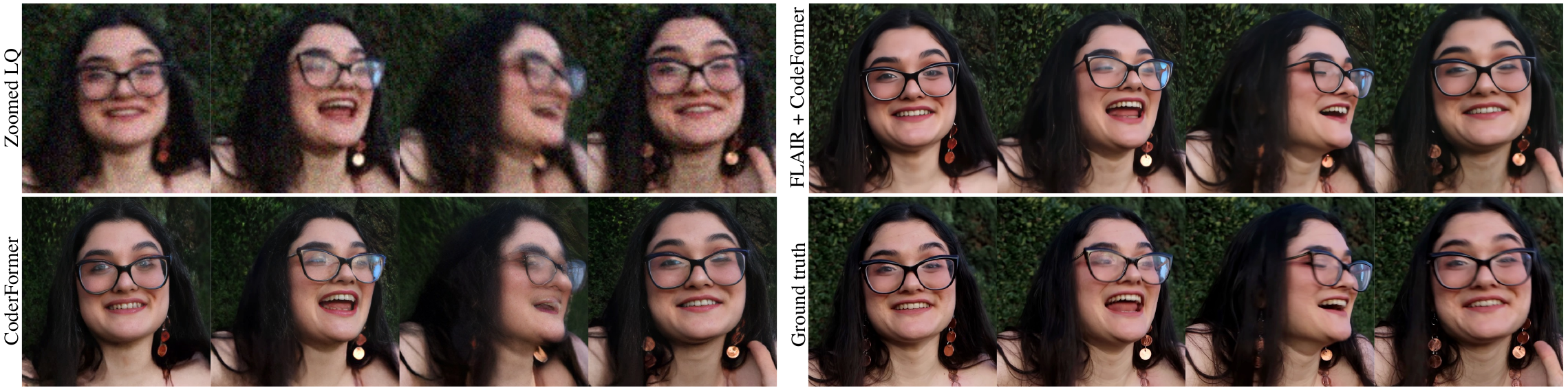}
    \vspace{-1em}
	\caption{Qualitative comparisons on face video motion deblurring. Thanks to our proposed video diffusion model, FLAIR produces high-quality and temporally consistent results than SOTA method~\cite{Zhou.etal2022towards}, even in the presence of large motion.}
	\label{Fig:fig9}
 \vspace{-1em}
\end{figure*}
\begin{figure}[t!]
	\centering
	\includegraphics[width=0.45\textwidth]{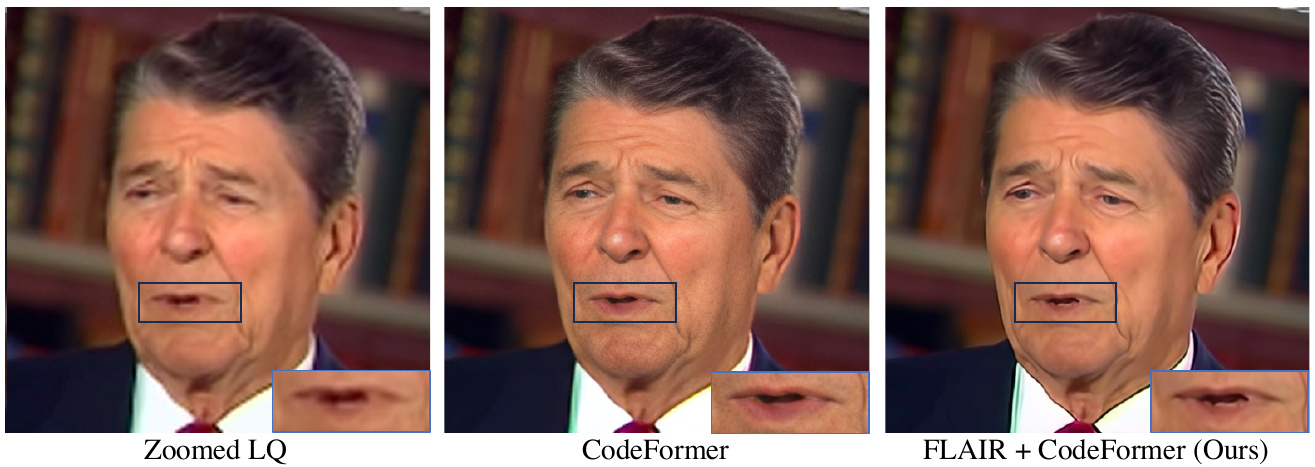}
    \vspace{-0.5em}
	\caption{Visual comparisons on~\emph{real-world} low-quality FVR. Note that FLAIR provides both high-quality and data-consistent result.}
	\label{Fig:fig10}
\vspace{-1.em}
\end{figure}
The quantitative results on short video clips from CelebV-Text and CelebV-HQ  are listed in Table~\ref{tab:table1}. As shown in the first two rows, FLAIR achieves the best performance on all evaluation metrics for both $16\times$ and $8\times$ upsampling tasks even~\emph{without} using any enhancement module, which is significant considering the severe degradation caused by low resolution. Despite DDNM obtains slightly higher PSNR and SSIM for 4$\times$ SR using the isotropic Gaussian kernel with a width of 2.0, FLAIR obtains the best LPIPS, FID and FVD scores, as shown in the third row. On the other hand, even with a more complex degradation ($4\times$ SR, Gaussian blur, AWGN $\sigma=0.05$, JPEG $Q=60$), our method continue to obtain superior scores across all metrics, showing our outputs have closer distribution to ground truth.

The quantitative results on long video clips from CelebV-Text are listed in Table~\ref{tab:table2}. We additionally compare two video restoration methods: VRT~\cite{Liang.etal2022vrt} and BasicVSRPP~\cite{Chan.etal2022basicvsr++}. VRT is a supervised deep learning approach to video SR and deblurring, while BasicVSRPP is a deep recurrent network method specifically designed for video SR. For motion deblurring task, we generate 100 distinct motion blur kernels using the methods in~\cite{Boracchi.etal2012modeling, Levin.etal2009understanding}. Each kernel is then applied to a frame by using~\eqref{eq:complexFVR} without JPEG encoding. Overall, our FLAIR + CodeFormer and FLAIR + RestoreFormer++ achieves the best LPIPS, FID and FVD scores thanks to the
pre-trained high-quality codebook. This suggests that our results are perceptually closer to the ground truth. We include FLAIR-SA (sampling average) to illustrate that different samples generated by our method achieve pixel-wise consistency in performance. Visual comparisons on single frame are presented in Fig.~\ref{Fig:fig4}. FLAIR produces fewer artifacts and more natural results on severely degraded inputs compared with previous methods.  In Fig~\ref{Fig:fig9}, we present the visual results on video motion deblurring. FLAIR provides more temporally aligned results, thanks to our proposed video DPM. By carefully constructing the forward model in~\eqref{eq:complexFVR}, one may directly applying FLAIR for real-world FVR (see supplementary material for more details), as shown in Fig~\ref{Fig:fig10}. 
\subsection{Ablation Studies}
\label{Sec:ablations}
\noindent
\textbf{Effect of Enhancement Module.}
We report PSNR and LPIPS results for our method in Fig.~\ref{Fig:fig6} by adjusting the weighted schedule $\{w_t\}_{t=\tau}^{K-1}, \tau\in[0,K-1]$ in~\eqref{eq:enhance}. For simplicity, we have selected RestoreFormer++ for $4\times$ SR with Gaussian blur kernel, and CodeFormer~\cite{Zhou.etal2022towards} for $16\times$ SR and JPEG $Q=60$. We consider growth sequences $1\geq w_{\tau}>\cdots > w_{K-1}=0$ from $K-1$ to $\tau$ and $w_t=0$ for $t<\tau$. Note that $w$ adjusts the relative weights of the enhancement module at each intermediate step. By setting an appropriate $w_{\tau}$, one can achieve perception (LPIPS) improvement for all three video tasks, with a very slight compromise on PSNR performance. Qualitative results are illustrated in Fig.~\ref{Fig:fig5}, demonstrating that FLAIR w/ the enhancement module yields superior visual outcomes.

\noindent
\textbf{Effect of Temporal Layers.}  In Table~\ref{tab:ewarp}, we show that FLAIR with video DPMs outperforms its image DPM counterpart in temporal consistency for video restoration. The temporal consistency is measured based on the averaged flow warping error $\mathrm{E}_\text{warp}(\xbm)=\frac{1}{N-1}\sum_{n=1}^{N-1}\mathrm{E}_\text{warp}(\xbm^n,\xbm^{n+1})$ over the entire sequence, as used by~\cite{Lai.etal2018learning, Lei.etal2020blind, Xu2022.etaltemporally, Lei.etal2023blind}, where lower value corresponds to smoother temporal results. Our temporal layers improve the sequential consistency of the restoration, outperforming SOTA video restoration method, VRT. 

\subsection{Space-Time Video Super-Resolution}
\label{Sec:space-time}
We show that the pre-trained FLAIR on video SR can be combined with any video frame interpolation method for space-time video SR. Here, we consider pre-trained AMT~\cite{li2023amt} for frame interpolation. In practice, we can cascade FLAIR in two ways: AMT followed by FLAIR, or FLAIR followed by AMT. As shown in Table~\ref{tab:st-sr}, compared with existing methods, FLAIR provides the best LPIPS, FVD, FID and KID scores, even though it serves as a two-stage model and is not specifically trained for this task. Additional details can be found in the supplements.

\begin{table}[t!]
\centering
\resizebox{\columnwidth}{!}{%
\begin{tabular}{lccccc}
\hline
         & CodeFormer~\cite{Zhou.etal2022towards} & VRT~\cite{Liang.etal2022vrt}   & \makecell{FLAIR \\(Image DPM)} & \makecell{FLAIR \\(Video DPM)} &  \makecell{FLAIR + CodeFormer \\(Video DPM)} \\ \hline\hline
$\mathrm{E}_{\text{warp}}\downarrow~(\times 10^{-3})$ & 3.928      & 2.639 & 5.625   & 2.546   & 2.531      \\ \hline
\end{tabular}%
}
\caption{Temporal inconsistency measured by warping error $\mathrm{E}_{\text{warp}}$, lower value corresponding to smoother temporal results.}
\label{tab:ewarp}
\vspace{-1em}
\end{table}
\begin{table}[t!]
\centering
\resizebox{\columnwidth}{!}{%
\begin{tabular}{lcccccc}
\hline
Method         & PSNR$\uparrow$  & SSIM$\uparrow$  & LPIPS$\downarrow$  & FVD$\downarrow$    & FID$\downarrow$    & KID$\downarrow$   \\ \hline\hline
\cite{li2023amt}+VRT~\cite{Liang.etal2022vrt}        & \colorbox{lightblue}{\makebox(24,4){30.03}} & 0.884 & 0.259 & 509.92 & 90.52  & 42.84 \\
VRT~\cite{Liang.etal2022vrt}+\cite{li2023amt}        & \colorbox{lightgreen}{\makebox(24,4){\bf30.87}} & \colorbox{lightgreen}{\makebox(24,4){\bf0.904}} & 0.190 & 275.78 & 62.75  & 27.92 \\
\cite{li2023amt}+DDNM~\cite{Wang.etal2023zeroshot}       & 28.93 & 0.863 & 0.266 & 435.32 & 94.87  & 52.76 \\
DDNM~\cite{Wang.etal2023zeroshot}+\cite{li2023amt}       & 29.19 & 0.872 & 0.247 & 329.15 & 91.06  & 52.84 \\
\cite{li2023amt}+VQFR~\cite{Gu.etal2022vqfr}       & 26.19 & 0.799 & 0.248 & 487.78 & 111.00    & 35.71 \\
VQFR~\cite{Gu.etal2022vqfr}+\cite{li2023amt}       & 26.35 & 0.816 & 0.251 & 473.47 & 111.51 & 35.93 \\
\cite{li2023amt}+DR2E~\cite{Wang.etal2023dr2}        & 27.08 & 0.823 & 0.218  & 533.37 & 81.38  & 25.36 \\
DR2E~\cite{Wang.etal2023dr2}+\cite{li2023amt}        & 27.34 & 0.841 & 0.220 & 470.57 & 83.47  & 26.78 \\
\cite{li2023amt}+Codeformer~\cite{Zhou.etal2022towards} & 26.52 & 0.801 & 0.246 & 535.59 & 105.60  & 35.68 \\
Codeformer~\cite{Zhou.etal2022towards}+\cite{li2023amt} & 26.67 & 0.819 & 0.244 & 596.61 & 104.53 & 35.09 \\
\cite{li2023amt}+Restoreformer++~\cite{Wang2023.etalrestoreformer++} & 26.68 & 0.805 & 0.236 & 415.03 & 100.91  & 32.49 \\
Restoreformer++~\cite{Wang2023.etalrestoreformer++}+\cite{li2023amt} & 26.85 & 0.822 & 0.237 & 477.78 & 99.05 & 31.54 \\ \hdashline
FLAIR (Ours)+\cite{li2023amt}       & 29.74 & \colorbox{lightblue}{\makebox(24,4){0.885}} & \colorbox{lightblue}{\makebox(24,4){0.182}} & \colorbox{lightblue}{\makebox(24,4){271.80}}  & \colorbox{lightblue}{\makebox(24,4){57.70}}   & \colorbox{lightblue}{\makebox(24,4){18.43}}\\
\cite{li2023amt} + FLAIR (Ours)     & 29.04 & 0.866 & \colorbox{lightgreen}{\makebox(24,4){\bf0.160}} & \colorbox{lightgreen}{\makebox(24,4){\bf242.60}}  & \colorbox{lightgreen}{\makebox(24,4){\bf51.42}}  & \colorbox{lightgreen}{\makebox(24,4){\bf12.24}} \\ \hline
\end{tabular}%
}
\caption{Quantitative results for space-time video super-resolution (time: $ 4\times$, space: $ 8\times$)~on CelebV-Text~\cite{Yu2023.etalcelebv} (long clips). AMT~\cite{li2023amt} is a SOTA frame interpolation method. Note that our FLIAR is only trained on spatial $8\times$ SR task. \colorbox{lightgreen}{\makebox(20,5.5){\textbf{Best}}}~and~\colorbox{lightblue}{\makebox(40,5.5){second-best}} values for each metric are color-coded.}
\vspace{-1em}
\label{tab:st-sr}
\end{table}
\section{Conclusion}
\label{Sec:conclusion}
In this paper, we propose the FLAIR, a novel framework based on diffusion probabilistic models for~\emph{face video restoration}. The key idea of FLAIR is to build upon pre-trained image diffusion models specialized in face image restoration and to transform them into video diffusion restoration models by incorporating and fine-tuning temporal alignment layers. We further propose a two-stage refinement process at every reverse sampling step. In the first stage, FLAIR analytically imposes reconstruction fidelity by using a data-consistency module that can handle composed degradation in practice. The subsequent stage involves an enhancement module dedicated to regional improvement. Extensive comparisons show that our FLAIR framework provides temporally aligned, high-quality results in face video restoration.
\section{Acknowledgements}
This paper is partially based upon work supported by the NSF CAREER award under grants CCF-2043134.

\newpage
\appendix
{\Large\textbf{~~~~~~Supplementary Material}}
\vspace{1em}
\medskip

\section{Additional Implementation Details}
In this section, we present additional implementation details omitted from the main paper due to space constraints. We train and evaluate all models with Pytorch on a computing cluster equipped with A40-40GB and A100-80GB GPUs. The detailed parameters setting is presented in Table~\ref{tab:hparam}.

\subsection{Training of conditional Image DPMs}
\begin{figure}[b!]
	\centering
	\includegraphics[width=0.47\textwidth]{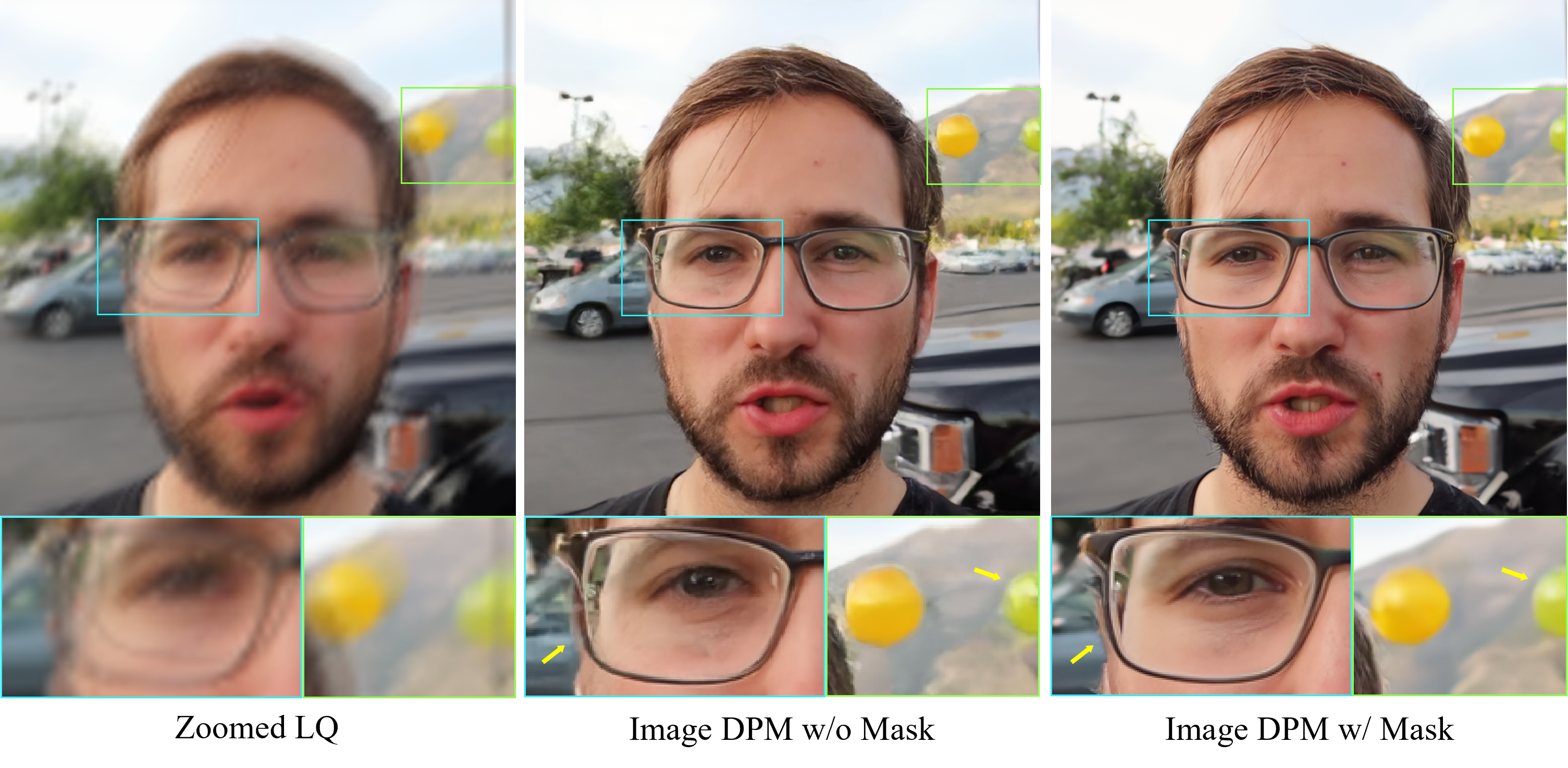}
    \vspace{-1.0em}
	\caption{Visual illustration of the impact of equation~\eqref{eq:mcondition} on training image DPMs. The zoomed-in regions are shown below the main results. Notably, the image restoration quality is improved by applying data augmentation to the conditional inputs. }
	\label{Fig:figwmask}
\vspace{-1.em}
\end{figure}

In order to improve the generation flexibility and empirical performance of FLAIR, we jointly train a single image diffusion model on conditional and unconditional objectives by randomly dropping $\cbm$ during training (\eg, $p_{\text{uncond}}=0.2$), similar to the \emph{classifier free guidance}~\cite{Ho.etal2021, Saharia.etal2022b}. Hence, the sampling is performed using the adjusted noise prediction:
\begin{equation}
    \label{eq:reverse4}
    \tilde\epsilon_{\thetabm}(\xbm_t,\cbm,t)=\lambda\epsilon_{\thetabm}(\xbm_t,\cbm,t) + (1-\lambda)\epsilon_{\thetabm}(\xbm_t,t),
\end{equation}where $\lambda>0$ is the trade-off parameter, and $\epsilon_{\thetabm}(\xbm_t,t)$ is the unconditional $\epsilonbm$-prediction. For example, setting $\lambda=1$ disables the unconditional guidance, while
increasing $\lambda > 1$ strengthens the effect of conditional $\epsilonbm$-prediction. 

Given that our video diffusion restoration models are fine-tuned on pre-trained image DPMs, it is reasonable to assume that a superior pre-trained image DPM would result in an better video DPM in terms of restoration quality. To this end, a data augmentation for training conditional image DPMs is done by constructing the conditional inputs $\cbm\in\R^{Nd}$ as follows
\begin{equation}
    \label{eq:mcondition}
    \cbm = \bm{m}_c \odot (\ybm)\uparrow^{s}_{bicubic},
\end{equation}
where $\bm{m}_c$ is a weighted mask that randomly reduces the importance of some pixels, analog to the masked augmentation training proposed in~\cite{He.etal2022masked}. We have observed that this data augmentation on $\cbm$ can improve the restoration results especially on large motion degradation, as shown in Fig.~\ref{Fig:figwmask}. The conditional input $\cbm$ is normalized to intensity range of $[-1,1]$ for better performance and stable training. We train all image DPMs in half precision (\texttt{float16}) with a batch-size of 64. We use the Adam optimizer with a fixed learning rate of $1.5\times 10^{-4}$ and a dropout rate of 0.2 for each model. In Fig.~\ref{Fig:figkernels}, we present samples of synthetically generated random kernels, following~\cite{zhang.etal2020a, Boracchi.etal2012modeling}, used to generate the image and video deblurring dataset.

\begin{figure}[t!]
    \centering
    \includegraphics[width=0.96\columnwidth]{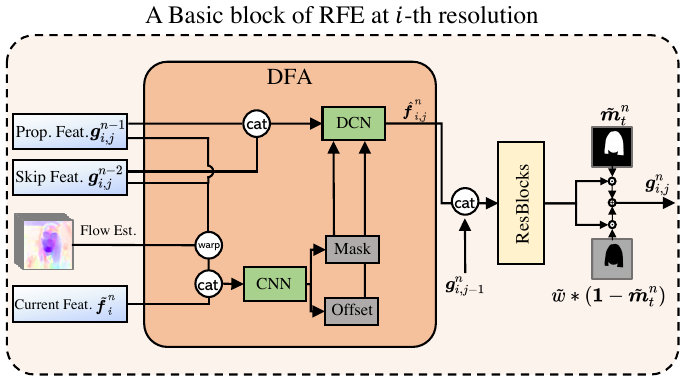}
    \caption{Illustration of one basic block of our proposed recurrent feature enhancement (RFE) module. The $\texttt{cat}$ operator denotes feature concatenation. }
    \label{fig:dcn}
    \vspace{-1em}
\end{figure}

\subsection{Implementations of Video DPM}
We use \texttt{einops}~\cite{rogozhnikov2021einops} to efficiently rearrange the features between spatial and temporal layers.

\noindent
\textbf{Group Normalization for Sequential Features.} For video DPMs, we observe that directly calculating group normalization to video features as independent images by rearranging the input as $\mathbb{R}^{B\times N\times C\times H\times W}\rightarrow\mathbb{R}^{(BN)\times C\times H\times W}$ results in temperature unalignment across frames. When calculating the group normalization, we consider the entire video by rearranging the input from $\mathbb{R}^{B\times N\times C\times H\times W}$ to $\mathbb{R}^{B\times C\times N\times H\times W}$, Consequently, the group normalization is computed along the $N$, $H$, $W$ axis. We have observed that applying this rearrangement to group normalization layers, which are pre-trained in image DPM, does not result in any performance degradation.

\begin{figure}[t!]
	\centering
	\includegraphics[width=\columnwidth]{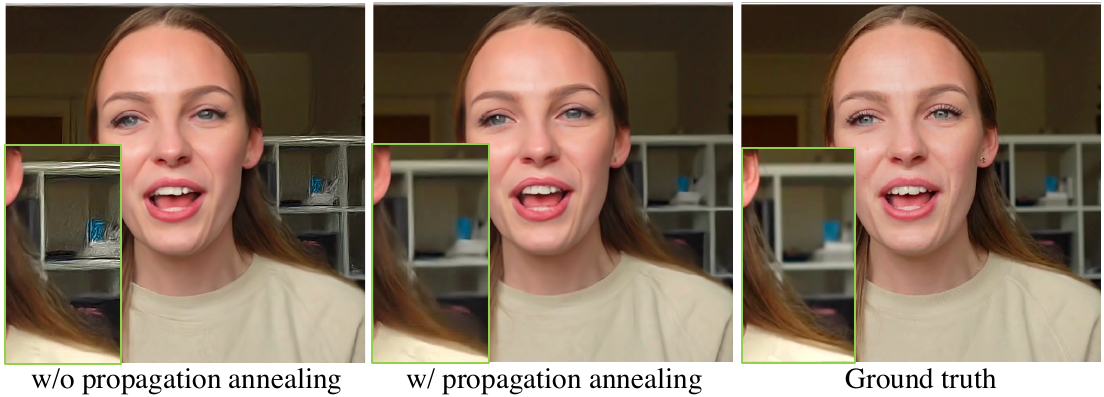}
    \vspace{-1.0em}
	\caption{Visual demonstration of the impact of our propagation annealing in~\eqref{eq:annealing} on $8\times$ SR task. The background scenes in each frame are improved, as shown in the zoomed-in figure.}
	\label{Fig:anneal}
\vspace{-0.em}
\end{figure}

\noindent
\textbf{More details about RFE Module.}
As introduced in the main paper, we implement recurrent feature enhancement (RFE) module to capture sequential dependencies and synchronize video frame features at high resolutions (\emph{e.g.,} [512, 256]). Fig~\ref{fig:dcn} illustrates one basic block of our RFE module. Given the extracted temporal features $\{\tilde\fbm_i^{n}\}_{n=1}^{N}$ from the 3D residual blocks at $i$-th resolution scale, we apply Deformable Feature Alignment (DFA)~\cite{Chan.etal2022basicvsr++} to propagate and align the intermediate features $\widehat\fbm_{i,j}^{n}$ as 
\begin{align*}
    &\hat\fbm_{i,j}^{n} = \texttt{DFA}(\tilde\fbm_i^{n}, \gbm_{i,j}^{n-1}, \gbm_{i,j}^{n-2}, \obm_i^{n\rightarrow n-1}, \obm_i^{n\rightarrow n-2}),
\end{align*} 
where $\gbm^{n-1}_{i,j}$ and $\gbm^{n-2}_{i,j}$ are the features at the $(n-1)$-th and $(n-2)$-th sequential step in the $j$-th propagation branch, respectively. For example, we have $\gbm^{n}_{i,0}=\tilde\fbm_i^{n}$.  Similarly, the $\obm_i^{n_1\rightarrow n_2}$ denotes the optical flow estimated from $n_1$-th degraded input frame to the $n_2$-th counterparts. The features $\hat\fbm_{i,j}^{n}$ are then concatenated (\texttt{cat}) and passed into a stack of residual blocks (\texttt{ResBlocks}) to fuse $\gbm_{i,j}^{n}$, denoted as
\begin{align}
    \label{eq:annealing}
    &\tilde\gbm_{i,j}^{n} = \hat\fbm_{i,j}^{n} + \texttt{ResBlocks}(\texttt{cat}(\gbm_{i,j-1}^{n},\hat\fbm_{i,j}^{n})),\\
    &\gbm_{i,j}^{n} = \tilde w*(\bm{1} - \tilde{\bm{m}}_t^n) \odot \tilde\gbm_{i,j}^{n} + (\tilde{\bm{m}}_t^n) \odot \tilde\gbm_{i,j}^{n},
\end{align}
where $\tilde w \in [0,1]$ balances the smoothness of the background scenes of the fused featur, denoted as $(\bm{1}-\tilde{\bm{m}}_t^n)\odot\tilde\gbm_{i,j}^n$. The masks $\{\tilde{\bm{m}}_t^n\}_{n=1}^N$ are the downscale version of facial region masks $\bm{m}_t = \{\bm{m}_t^n\}_{n=1}^N$ estimated from $\xbm_{0t}$ at the $t$-th reverse diffusion step. The main motivation behind the design of propagation annealing is to enhance robustness against appearance changes and error accumulation within the recurrent network. We have observed that this annealing can notably improve the temporal consistency of background scenes across frames while preserving the sharpness of facial region, as shown in Fig~\ref{Fig:anneal}. 

\subsection{Training of video DPMs}
All video DPMs are fine tuned with batch size $B=4$ and frame length $N=10$. 
We set schedule $T=1000$ and uniformly spaced $\beta_t$ for both video deblurring and JPEG restoration, while $T=2000$ for video super-resolution tasks. We use the Adam optimizer with a fixed learning rate of $1\times 10^{-4}$ and weight-decay of $0.05$ for fine-tuning the video DPMs. Similarly, we train all DPMs in half precision ($\texttt{float16}$). We do not apply classifier free guidance for fine-tuning video diffusion model. Note that, we do not perform any checkpoint selection on our models and simply select the latest checkpoint of each model. It will take around a week to get a video DPM. 

\subsection{Implementations during Inference}
Our proposed reverse diffusion sampling is illustrated in Algorithm~\ref{alg:redbls}. We use an exponential decay for $\gamma_t$, where we parameterize $\gamma_t=1-\zeta\frac{\sigma^2_\ebm\bar\alpha_{t
}}{\bar\alpha_{t-1}}$, where $\zeta$ controls the strength of the data consistency module, and $\gamma$ is clipped into range $[0, 1]$. The setting of $\zeta$ for each task is presented in Table \ref{tab:hparam}. We use an exponential growth for $\{w_t\}_{t=\tau}^{K-1}$. We parameterize $w_t =e^{-(t-\tau)/(K-\tau)} * w_\tau$, where $w_\tau$ controls the final strength of the enhancement module, and $\tau$ controls where the enhancement modules end its participation during sampling. The setting of $w_\tau$ and $\tau$ for each task can be found in Table \ref{tab:hparam}. We run a grid search for best controlling hyperparameters of the two-stage conditional refinement and the rescheduling time step $K$ for each dataset, similar to~\cite{Wang.etal2023dr2, Whang.etal2022, Wang.etal2023zeroshot, Zhu.etal2023denoising}. 
This inference-time hyperparameter tuning is cheap as it does not involve retraining or fine-tuning the model itself. The facial mask $\bm{m}_t$ estimation follows the similar method as~\cite{Yang.etal2021gan, Zhou.etal2022towards, Gu.etal2022vqfr, Wang.etal2023dr2}, where we introduce in a separate subsection~\ref{Sec:image_processing}.

\begin{algorithm}[t]
\caption{FLAIR Face Video Iterative Refinement}\label{alg:redbls}
\begin{algorithmic}[1]
\State \textbf{Input:} $\epsilon_{\thetabm,\phibm}$: Video denoiser network; $\ybm$: Degraded video;
$\mathcal{G}:$ Image Enhancement module; $\gamma_t, \rho_t, w_t$;
\State \textbf{Output:} Restored video $\xbm_0$
\State \text{Sample} $\xbm_{T} \sim \Ncal({\zerobm, \Ibf})$   \Comment{Run diffusion sampling}
\For{$t = T,\dots,1$}
\State $\epsilonbm\sim\Ncal(\zerobm,\Ibf)$
\State $\xbm_{0t}=\frac{1}{\sqrt{\bar\alpha_t}}(\xbm_t + (1- \Bar{\alpha}_t)\epsilon_{\thetabm,\phibm}(\xbm_t,\cbm,t))$
\State $\tilde\xbm_{0t}=\xbm_{0t} -\gamma_t(\Acal^{+}\Acal\xbm_{0t} -  \Acal^{+}\ybm)$
\State $\tilde\xbm_{0t}= (\bm{1}-w_t \bm{m}_t)\odot\tilde\xbm_{0t} + w_t \bm{m}_t\odot\mathcal{G}(\tilde\xbm_{0t})$
\State $\tilde\epsilonbm_t=\frac{1}{\sqrt{1-\bar\alpha_t}}(\xbm_t - \sqrt{\bar\alpha_t}\tilde\xbm_{0t})$
\State $\xbm_{t-1}=\sqrt{\bar\alpha_{t-1}}\tilde\xbm_{0t} + \sqrt{1-\bar\alpha_t}(\sqrt{1-\rho_t}\tilde\epsilonbm_t + \sqrt{\rho_t}\epsilonbm)$
\EndFor
\State \textbf{return:} $\xbm_0$
\end{algorithmic}
\end{algorithm}
\begin{figure}[t!]
	\centering
	\includegraphics[width=0.47\textwidth]{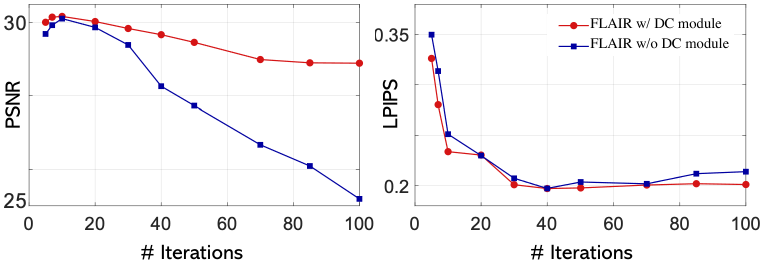}
    \vspace{-1.0em}
	\caption{Comparison of average PSNR (left) and LPIPS (right) of FLAIR w/ and w/o data-consistency module for FVR with a mixed degradation ($4\times$ SR, Gaussian blur width$=2$, $\sigma=0.05$, JPEG $Q=60$). Both methods use uniform re-scheduling strategy starting from $K=5$ to $K=100$. Note the improved data-fidelity (PSNR) by imposing data-consistency and the trade-off between perception and distortion~\cite{Blau.etal2018perception} during inference. 
 }
	\label{Fig:figdcablation}
\vspace{-1.em}
\end{figure}
\begin{table*}[]
\centering
\resizebox{0.93\textwidth}{!}{%
\begin{tabular}{l|c|cccccc|cccccc}
\hline
\multirow{2}{*}{Method} & \multirow{2}{*}{Task} & \multicolumn{6}{c}{CelebV-Text~\cite{Yu2023.etalcelebv}}                       & \multicolumn{6}{c}{CelebV-HQ~\cite{Zhu.etal2022celebv}}                         \\ \cline{3-14} 
                        &                       & PSNR  & SSIM   & LPIPS  & FVD    & FID   & KID   & PSNR  & SSIM   & LPIPS  & FVD    & FID   & KID   \\ \hline
VQFR~\cite{Gu.etal2022vqfr}                    & \multirow{5}{*}{\rotatebox{90}{$8\times$ Bicubic}}     & 26.34 & 0.805 & 0.221 & 238.89 & 46.53 & 9.92  & 26.37 & 0.793 & 0.219 & 528.02 & 74.01 & 14.76 \\
CodeFormer~\cite{Zhou.etal2022towards}              &                       & 26.60  & 0.783  & 0.238  & 215.07 & 50.03 & 12.40  & 26.64 & 0.770 & 0.236 & 444.52 & 81.58 & 20.44 \\
RestoreFormer++~\cite{Wang2023.etalrestoreformer++}         &                       & \colorbox{lightblue}{\makebox(24,4){27.13}} & 0.792  & 0.225 & \colorbox{lightblue}{\makebox(24,4){130.64}} & \colorbox{lightblue}{\makebox(24,4){42.64}} & \colorbox{lightblue}{\makebox(24,4){8.58}}  & \colorbox{lightblue}{\makebox(24,4){27.69}}& 0.790 & \colorbox{lightblue}{\makebox(24,4){0.208}} & \colorbox{lightblue}{\makebox(24,4){330.02}} & \colorbox{lightblue}{\makebox(24,4){61.94}} & \colorbox{lightblue}{\makebox(24,4){14.26}} \\
DR2E~\cite{Wang.etal2023dr2}                   &                       & 26.59 & \colorbox{lightblue}{\makebox(24,4){0.810}} & \colorbox{lightblue}{\makebox(24,4){0.220}} & 243.15 & 46.62 & 10.95 & 26.56 & \colorbox{lightblue}{\makebox(24,4){0.798}} & 0.216 & 556.67 & 73.16 & 15.22 \\
FLAIR (Ours)                    &                       & \colorbox{lightgreen}{\makebox(24,4){\bf32.13}} & \colorbox{lightgreen}{\makebox(24,4){\bf0.889}} & \colorbox{lightgreen}{\makebox(24,4){\bf0.139}}  & \colorbox{lightgreen}{\makebox(24,4){\bf62.43}}  & \colorbox{lightgreen}{\makebox(24,4){\bf31.93}} & \colorbox{lightgreen}{\makebox(24,4){\bf6.29}}  & \colorbox{lightgreen}{\makebox(24,4){\bf31.80}}  & \colorbox{lightgreen}{\makebox(24,4){\bf0.875}} & \colorbox{lightgreen}{\makebox(24,4){\bf0.132}} & \colorbox{lightgreen}{\makebox(24,4){\bf146.57}} & \colorbox{lightgreen}{\makebox(24,4){\bf42.06}} & \colorbox{lightgreen}{\makebox(24,4){\bf6.68}}  \\ \hline
VQFR~\cite{Gu.etal2022vqfr}                    & \multirow{5}{*}{\rotatebox{90}{$16\times$ Bicubic}}     & 24.31 & \colorbox{lightblue}{\makebox(24,4){0.762}} & 0.270 & 383.47 & 55.04 & 13.69 & 24.28 & \colorbox{lightblue}{\makebox(24,4){0.743}} & 0.268 & 797.95 & 88.40  & 19.94 \\
CodeFormer~\cite{Zhou.etal2022towards}              &                       & \colorbox{lightblue}{\makebox(24,4){24.39}} & 0.732 & 0.298 & 397.34 & 59.57 & 16.20  & \colorbox{lightblue}{\makebox(24,4){24.37}} & 0.713 & 0.302 & 865.36 & 98.22 & 25.64 \\
RestoreFormer++~\cite{Wang2023.etalrestoreformer++}         &                       & 23.70  & 0.719 & 0.295 & \colorbox{lightblue}{\makebox(24,4){284.66}} & 56.20  & \colorbox{lightblue}{\makebox(24,4){12.17}} & 24.36 & 0.715 & 0.279 & \colorbox{lightblue}{\makebox(24,4){615.80}}  & 89.85 & 19.77 \\
DR2E~\cite{Wang.etal2023dr2}                    &                       & 24.23 & 0.755 & 0.271 & 400.64 & \colorbox{lightblue}{\makebox(24,4){51.95}} & 12.45 & 24.33 & 0.741 & \colorbox{lightblue}{\makebox(24,4){0.266}} & 722.86 & \colorbox{lightblue}{\makebox(24,4){84.81}} & \colorbox{lightblue}{\makebox(24,4){17.62}} \\
FLAIR (Ours)                    &                       & \colorbox{lightgreen}{\makebox(24,4){\bf28.49}}  & \colorbox{lightgreen}{\makebox(24,4){\bf0.844}} & \colorbox{lightgreen}{\makebox(24,4){\bf0.230}} & \colorbox{lightgreen}{\makebox(24,4){\bf201.86}} & \colorbox{lightgreen}{\makebox(24,4){\bf50.73}} & \colorbox{lightgreen}{\makebox(24,4){\bf10.24}} & \colorbox{lightgreen}{\makebox(24,4){\bf28.31}} & \colorbox{lightgreen}{\makebox(24,4){\bf 0.808}} & \colorbox{lightgreen}{\makebox(24,4){\bf0.216}} & \colorbox{lightgreen}{\makebox(24,4){\bf413.81}} & \colorbox{lightgreen}{\makebox(24,4){\bf78.38}} & \colorbox{lightgreen}{\makebox(24,4){\bf11.68}} \\ \hline
\end{tabular}%
}
\caption{Quantitative results calculated only within face regions on two video datasets (short clips). VQFR, CodeFormer, RestoreFormer++ and DR2E are SOTA face restoration methods that rely on separate methods for backgrounds enhancement. Note the quantitative improvements achieved by FLAIR  when it is specifically evaluated on face regions. \colorbox{lightgreen}{\makebox(20,5.5){\textbf{Best}}}~and~\colorbox{lightblue}{\makebox(40,5.5){second-best}} values for each metric are color-coded.}
\label{tab:only_face}
\end{table*}

\begin{table*}[]
\centering
\resizebox{0.93\textwidth}{!}{%
\begin{tabular}{l|cccccc|cccccc}
\hline
\multirow{2}{*}{Method}      & \multicolumn{6}{c}{CelebV-Text~\cite{Yu2023.etalcelebv}}                       & \multicolumn{6}{c}{CelebV-HQ~\cite{Zhu.etal2022celebv}}                         \\ \cline{2-13} 
                             & PSNR  & SSIM   & LPIPS  & FVD    & FID   & KID   & PSNR  & SSIM   & LPIPS  & FVD    & FID   & KID   \\ \hline 
VQFR~\cite{Gu.etal2022vqfr}                         & 28.88 & 0.855 & 0.160 & 151.86 & 46.25 & 10.34 & 28.59 & 0.847 & 0.156 & 261.27 & 66.98 & 14.50  \\
CodeFormer~\cite{Zhou.etal2022towards}                   & 29.80  & 0.867 & 0.153 & 107.39 & 45.46 & 10.6  & 29.17 & 0.856 & 0.151 & 219.77 & 66.42 & 15.53 \\
RestoreFormer++~\cite{Wang2023.etalrestoreformer++}              & 29.06 & 0.856 & 0.151  & 111.53 & 45.80 & 10.21 & 28.96 & 0.849 & 0.149 & 211.02 & 65.51 & 12.60 \\
DR2E~\cite{Wang.etal2023dr2}                          & 28.40  & 0.836 & 0.167 & 189.91 & 44.49 & 9.18  & 27.98 & 0.800 & 0.163 & 378.15 & 76.39 & 15.33 \\
DDNM~\cite{Wang.etal2023zeroshot}                         & 34.76 & 0.929 & 0.118 & 31.48  & 37.65 & 20.28 & 33.46 & 0.917 & 0.129 & 89.33  & 55.27 & 27.89 \\ \hdashline
FLAIR (Ours)                 & \colorbox{lightgreen}{\makebox(24,4){\bf36.05}} & \colorbox{lightgreen}{\makebox(24,4){\bf0.942}} & 0.061 & \colorbox{lightblue}{\makebox(24,4){26.57}}  & 11.27 & 2.64  & \colorbox{lightgreen}{\makebox(24,4){\bf34.46}} & \colorbox{lightgreen}{\makebox(24,4){\bf0.932}} & 0.060 & \colorbox{lightblue}{\makebox(24,4){76.18}}  & 15.36 & 1.50   \\
FLAIR+CodeFormer (Ours)      & 35.10  & 0.934 & \colorbox{lightblue}{\makebox(24,4){0.059}} & \colorbox{lightgreen}{\makebox(24,4){\bf26.44}}  & \colorbox{lightgreen}{\makebox(24,4){\bf9.51}}  & \colorbox{lightgreen}{\makebox(24,4){\bf0.75}}  & 33.47 & 0.920 & \colorbox{lightblue}{\makebox(24,4){0.059}} & \colorbox{lightgreen}{\makebox(24,4){\bf74.56}}  & \colorbox{lightgreen}{\makebox(24,4){\bf13.84}} & \colorbox{lightgreen}{\makebox(24,4){\bf0.04}}  \\
FLAIR+RestoreFormer++ (Ours) & \colorbox{lightblue}{\makebox(24,4){35.42}} & \colorbox{lightblue}{\makebox(24,4){0.936}} & \colorbox{lightgreen}{\makebox(24,4){\bf0.057}} & 27.22  & \colorbox{lightblue}{\makebox(24,4){10.24}} & \colorbox{lightblue}{\makebox(24,4){1.59 }}  & \colorbox{lightblue}{\makebox(24,4){34.17}} & \colorbox{lightblue}{\makebox(24,4){0.927}} & \colorbox{lightgreen}{\makebox(24,4){\bf0.056}} & 78.07  & \colorbox{lightblue}{\makebox(24,4){14.49}} & \colorbox{lightblue}{\makebox(24,4){0.69}}  \\ \hline
\end{tabular}%
}
\caption{Quantitative results of $4\times$ face video super-resolution on two separate video datasets (short clips). Note the quantitative improvements achieved by integrating our enhancement module within FLAIR, even in cases of mild degradation. \colorbox{lightgreen}{\makebox(20,5.5){\textbf{Best}}}~and~\colorbox{lightblue}{\makebox(40,5.5){second-best}} values for each metric are color-coded.}
\label{tab:x4_bicubic}
\end{table*}
\begin{table}[]
\centering
\resizebox{\columnwidth}{!}{%
\begin{tabular}{lcccccc}
\hline
Method         & PSNR$\uparrow$  & SSIM$\uparrow$   & LPIPS$\downarrow$  & FVD$\downarrow$    & FID$\downarrow$   & KID$\downarrow$   \\ \hline\hline
\cite{li2023amt}+VRT~\cite{Liang.etal2022vrt}        & \colorbox{lightblue}{\makebox(24,4){33.10}} & \colorbox{lightblue}{\makebox(24,4){0.936}} & 0.112 & 194.57 & 35.78 & 15.00    \\
VRT~\cite{Liang.etal2022vrt}+~\cite{li2023amt}        & \colorbox{lightgreen}{\makebox(24,4){\bf33.47}} & \colorbox{lightgreen}{\makebox(24,4){\bf0.941}} & 0.085 & \colorbox{lightgreen}{\makebox(24,4){\bf177.89}} & \colorbox{lightblue}{\makebox(24,4){\colorbox{lightblue}{\makebox(24,4){20.65}}}} & 6.92  \\
\cite{li2023amt}+DDNM~\cite{Wang.etal2023zeroshot}       & 32.21 & 0.922 & 0.136 & 199.86 & 50.68 & 27.44 \\
DDNM~\cite{Wang.etal2023zeroshot}+~\cite{li2023amt}       & 29.52 & 0.873  & 0.170 & 194.38 & 50.65 & 26.93 \\
\cite{li2023amt}+VQFR       & 27.90  & 0.844  & 0.166 & 385.31 & 62.68 & 18.55 \\
VQFR~\cite{Gu.etal2022vqfr}+~\cite{li2023amt}       & 27.84 & 0.855 & 0.172 & 368.36 & 61.07 & 17.94 \\
\cite{li2023amt}+DR2        & 27.57 & 0.834  & 0.175 & 457.47 & 57.58 & 15.10  \\
DR2E~\cite{Wang.etal2023dr2}+~\cite{li2023amt}        & 27.69 & 0.850 & 0.186 & 407.21 & 63.11 & 19.15 \\
\cite{li2023amt}+CodeFormer~\cite{Zhou.etal2022towards} & 29.13 & 0.860 & 0.151 & 334.95 & 55.45 & 18.11 \\
CodeFormer~\cite{Zhou.etal2022towards}+~\cite{li2023amt} & 29.06 & 0.872  & 0.151 & 342.28 & 53.38 & 17.56 \\
~\cite{li2023amt}+RestoreFormer++~\cite{Wang2023.etalrestoreformer++} & 29.36 & 0.864 & 0.147 & 307.01 & 54.64 & 17.95 \\
RestoreFormer++~\cite{Wang2023.etalrestoreformer++}+~\cite{li2023amt} & 29.55 & 0.883  & 0.148 & 312.84 & 52.26 & 17.01 \\\hdashline
FLAIR (Ours)+~\cite{li2023amt}       & 32.96 & 0.934 & \colorbox{lightblue}{\makebox(24,4){0.083}} & \colorbox{lightblue}{\makebox(24,4){179.38}} & 21.51 & \colorbox{lightblue}{\makebox(24,4){6.88}}  \\
\cite{li2023amt}+FLAIR (Ours)     & 32.49 & 0.929 & \colorbox{lightgreen}{\makebox(24,4){\bf0.077}} & 179.46 & \colorbox{lightgreen}{\makebox(24,4){\bf18.64}} & \colorbox{lightgreen}{\makebox(24,4){\bf4.66}}  \\ \hline
\end{tabular}%
}
\caption{Quantitative results of space-time video super-resolution (time: $ 4\times$, space: $ 4\times$)~on CelebV-Text~\cite{Yu2023.etalcelebv} (long clips). AMT~\cite{li2023amt} is a SOTA frame interpolation method. Note that our FLIAR is only trained on spatial $4\times$ SR task. \colorbox{lightgreen}{\makebox(20,5.5){\textbf{Best}}}~and~\colorbox{lightblue}{\makebox(40,5.5){second-best}} values for each metric are color-coded.}
\label{tab:x4-stsr}
\end{table}

\begin{table}[t!]
\centering
\resizebox{0.60\columnwidth}{!}{%
\begin{tabular}{lccc}
\hline
Method                  & PSNR  & SSIM  & LPIPS \\ \hline
VRT~\cite{Liang.etal2022vrt}                     & 31.24 & 0.911 & 0.140 \\
CodeFormer~\cite{Zhou.etal2022towards}              & 24.62 & 0.798 & 0.189 \\
RestoreFormer++~\cite{Wang2023.etalrestoreformer++}              & 24.58 & 0.796 & 0.180 \\
FLAIR (Ours)            & 31.48 & 0.902 & 0.085 \\\hline

\end{tabular}%
}
\caption{Quantitative results of $4\times$ super-resolution, motion deblurring with AWGN $\sigma=0.05$ on Obama dataset~\cite{Suwajanakorn.etal2017synthesizing}.}
\vspace{-1em}
\label{tab:obama}
\end{table}
\begin{table}[]
\centering
\resizebox{\columnwidth}{!}{%
\begin{tabular}{lcccccc}
\hline
Method                         & PSNR  & SSIM  & LPIPS & FVD    & FID   & KID   \\ \hline
\multicolumn{7}{c}{CelebV-Text~\cite{Yu2023.etalcelebv} (short clips)}                                                 \\ \hline
FLAIR (Ours)                   & 29.87 & 0.856 & 0.149 & 82.82  & 39.54 & 8.25  \\
FLAIR+Unconditional DPM (Ours) & 30.73 & 0.865 & 0.157 & 81.09  & 45.48 & 12.65 \\ \hline

\multicolumn{7}{c}{CelebV-Text~\cite{Yu2023.etalcelebv} (long clips)}                                    \\ \hline
FLAIR (Ours)                   & 31.51 & 0.858 & 0.169 & 175.52 & 55.88 & 20.85 \\
FLAIR+Unconditional DPM (Ours) & 31.44 & 0.859 & 0.163 & 146.31 & 55.69 & 20.95 \\ \hline
\end{tabular}%
}
\caption{Quantitative results of FLAIR using unconditional image DPM as enhancement module for $4\times$ super-resolution, Gaussian deblurring, AWGN $\sigma=0.05$ on CelebV-Text~\cite{Yu2023.etalcelebv}.}
\label{tab:unconditional}
\end{table}

\begin{table}[]
\centering
\resizebox{0.7\columnwidth}{!}{%
\begin{tabular}{lc}
\hline
Method                      & Sampling Time (sec)  \\\hline 
DDNM~\cite{Wang.etal2023zeroshot}                 & 42.95\\
FLAIR (Ours)               & 112.53  \\ 
FLAIR+CodeFormer (Ours)    & 137.43  \\
FLAIR+RestoreFormer (Ours) & 138.01  \\\hline
\end{tabular}%
}
\caption{Averaged runtime comparisons between FLAIR and other image DPM baselines for generating 10 frames. The experiments have been conducted on A100-80G for $4\times$ SR video JPEG restoration. }
\vspace{-1.5em}
\label{tab:runtime}
\end{table}

\begin{figure*}[t!]
	\centering
	\includegraphics[width=0.78\textwidth]{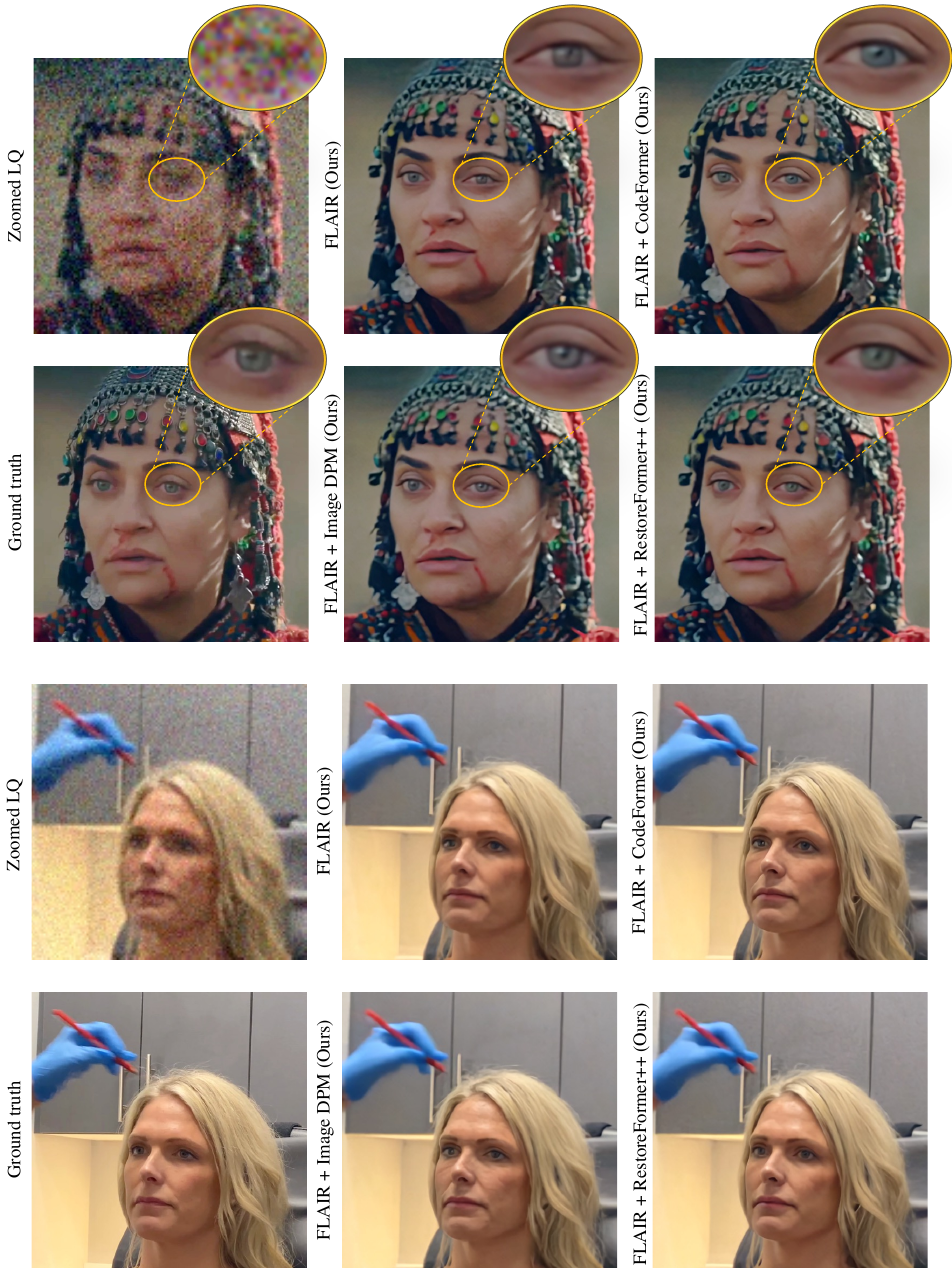}
    \vspace{-1em}
\caption{Visual comparisons of $4\times$ face video super-resolution with Gaussion blur kernel of width$ =2$ and AWGN $\sigma=0.05$ on CelebV-Text~\cite{Yu2023.etalcelebv} (\emph{top}) and CelebV-HQ~\cite{Zhu.etal2022celebv} (\emph{bottom}), respectively. Note the perceptual quality improvements of our FLAIR by applying different backbones for facial region enhancement. Best viewed by zooming in the display. }
	\label{Fig:visua5564}
\vspace{-1.em}
\end{figure*}
\begin{figure*}[t!]
	\centering
	\includegraphics[width=0.98\textwidth]{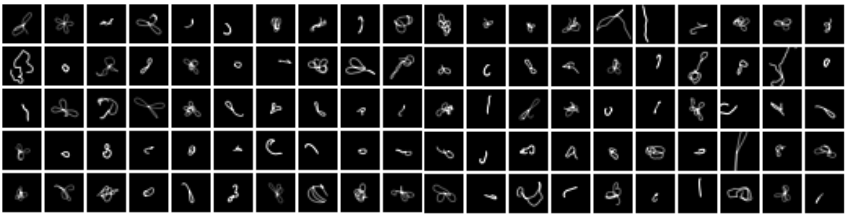}
    \vspace{-1.0em}
	\caption{Examples of synthetically generated random kernels, following~\cite{zhang.etal2020a, Boracchi.etal2012modeling}, used to generate the video deblurring dataset.}
	\label{Fig:figkernels}
\vspace{-0.em}
\end{figure*}

\subsection{Baseline Methods}
\noindent
\textbf{CodeFormer~\cite{Zhou.etal2022towards}, VQFR~\cite{Gu.etal2022vqfr} and RestoreFormer++~\cite{Wang2023.etalrestoreformer++}} refer to recently developed conditioning generative methods that use pre-trained Vector-Quantization (VQ) codebooks as dictionaries, achieving SOTA results in blind face restoration. These codebooks are learned on the entire facial region. We employ their original implementations~\footnote{\url{https://github.com/sczhou/CodeFormer}}$^{,}$\footnote{\url{https://github.com/TencentARC/VQFR}}$^{,}$\footnote{\tiny{\url{https://github.com/wzhouxiff/RestoreFormerPlusPlus}}} and pre-trained models for our tasks. For all these three baseline methods, we follow their original implementations of frame background enhancement accordingly.

\noindent
\textbf{VRT~\cite{Liang.etal2022vrt}} denotes a recently developed video restoration transformer (VRT) method, characterized by its parallel frame prediction and long-range temporal dependency modeling abilities. VRT has been shown superior performance for general restoration tasks such as video denoising, deblurring, super-resolution, etc. We modify the publicly available implementation~\footnote{\url{https://github.com/JingyunLiang/VRT}} and train the model for each task on the same CelebV-Text~\cite{Yu2023.etalcelebv} video training dataset as FLAIR.

\noindent
\textbf{BasicVSPP~\cite{Chan.etal2022basicvsr++}} is another recent SOTA method based on recurrent refinement structure for video super-resolution. BasicVSPP improves over BasicVSR~\cite{Chan.etal2021basicvsr} by proposing a second-order grid propagation with flow guided deformable alignment. Likewise, we modify the publicly available implementation~\footnote{\url{https://github.com/open-mmlab/mmagic}} and train the model on the same CelebV-Text~\cite{Yu2023.etalcelebv} training dataset as FLAIR.

\noindent
\textbf{ILVR~\cite{Choi.etal2021} and DR2E~\cite{Wang.etal2023dr2}} are two recently developed conditioning methods based on unconditionally trained
image DPM for solving versatile blind image restoration tasks. Both ILVR and DR2E share the similar conditional sampling implementation, whereas DR2E adapts an additional enhancement module for face regions similar to FLAIR. We modify the publicly available implementation~\footnote{{\url{https://github.com/jychoi118/ilvr_adm}}}$^{,}$\footnote{\tiny{\url{https://github.com/Kaldwin0106/DR2_Drgradation_Remover}}} of
both methods for each FVR task. We use the similar grid search to FLAIR for fine-tuning the hyper-parameters within
ILVR and DR2E, respectively.

\noindent
\textbf{DDNM~\cite{Choi.etal2021} and DiffPIR~\cite{Wang.etal2023dr2}} refer to recently developed conditioning methods based on unconditionally trained
image DPM for solving general image inverse problems. Unlike ILVR and DR2E, DDNM and DiffPIR rely on the forward-model to impose data-consistency. Similarly, we modify the publicly available implementation~\footnote{{\url{https://github.com/wyhuai/DDNM}}}$^{,}$\footnote{\tiny{\url{https://github.com/yuanzhi-zhu/DiffPIR}}} of
both methods for each FVR task. We use the similar grid search to FLAIR for fine-tuning the hyper-parameters within
DDNM and DiffPIR, respectively.

\noindent
We pre-train an unconditional image DPM on FFHQ and then fine tune it on the same CelebV-Text images used for video DPMs as additional baseline. All diffusion model based baseline methods, including ILVR, DR2E, DDNM, DiffPIR share the same unconditional image DPM. We train the baseline unconditional diffusion model modified based on the publicly available PyTorch implementation~\footnote{\url{https://github.com/openai/guided-diffusion}} for around $1\times 10^7$ samples in total (pre-training and fine-tuning).

\subsection{Face Detection and Processing}
\label{Sec:image_processing}
We process the images using the tools provided in \texttt{facexlib}\footnote{\url{https://github.com/xinntao/facexlib}}. 

\noindent
\textbf{Face Region Affine Transformation.} We first use \texttt{RetinaFace}~\footnote{\url{https://github.com/biubug6/Pytorch_Retinaface}} to calculate the face landmarks. Then we use \texttt{OpenCV}~\cite{opencv_library} to estimate affine matrices and transform the images to the head-only version with bicubic interpolation.

\noindent
\textbf{Estimation of Face Mask $\bm{m}_t$.}
We use \texttt{ParseNet}~\cite{Chen.etal2021semantic} to get the face parsing map, and convert it to a soft mask $\bm{m}_t$ with Gaussian blurring. The above process has been widely adapted for FVR in recent methods, such as~\cite{Yang.etal2021gan, Wang.etal2021towards, Zhou.etal2022towards, Gu.etal2022vqfr, Wang2023.etalrestoreformer++, Wang.etal2023dr2}.

\section{Datasets}

\noindent
\textbf{CelebV-HQ~\cite{Zhu.etal2022celebv}} dataset is a large-scale, high-quality video dataset with rich facial attributes for video generation and editing. CelebV-HQ contains $35, 666$ video clips with the
resolution of $512 \times 512$ at least. All data is publicly available~\footnote{\url{https://celebv-hq.github.io/}}. We randomly select $20$ clips, each containing $25$ high quality sequences from CelebV-HQ. 

\noindent
\textbf{CelebV-Text~\cite{Yu2023.etalcelebv}} dataset is another large-scale, high-quality, diverse dataset of facial text-video pairs. CelebV-Text comprises $70,000$ in-the-wild face video clips with diverse visual content. All data is publicly available~\footnote{\url{https://celebv-text.github.io/}}. we select $7200$ clips with each containing $20$ high quality $512\times512$ sequences for training. For video testing datasets, we randomly chose $125$ short clips and $6$ long clips from the unused portion of the CelebV-Text, ensuring no identity overlap with the fine-tuning datasets. Each short clip contains $25$ sequences, and each long clip contains $100$ sequences. As highlighted by its original authors, the videos that have appeared in CelebV-HQ are filtered out.

\noindent
\textbf{Obama Clip.} We select the video part C~\footnote{\tiny{\url{https://www.youtube.com/watch?v=deF-f0OqvQ4\&t=97s}}} from the Obama dataset~\cite{Suwajanakorn.etal2017synthesizing}. We extract the first $100$ frames from original videos. We crop out the head-only region from the frames using the same processes described in~\ref{Sec:image_processing}.

\noindent
\textbf{Web Video Clip.} We extract a low quality web video of $300$ frames from Internet~\footnote{\tiny{\url{https://www.youtube.com/watch?v=80vhQ1fypOU?vq=small}}}, which suffers from complex unknown degradation. The collected clip is then crop out the face-only region using the same processes as in \ref{Sec:image_processing}, following~\cite{Yang.etal2021gan, Zhou.etal2022towards, Gu.etal2022vqfr}.

\section{Additional Results}
We present additional experimental results that were omitted from the main paper due to space limitations.~\emph{We provide several video comparisons of our FLAIR in the supplementary materials.}

\subsection{Additional Numerical Results}
\noindent
\textbf{Numerical Evaluation on Facial Region Only.} Given that some of the state-of-the-art (SOTA) methods, including VQFR, CodeFormer, RestoreFormer++, and DR2E, are primarily designed for face restoration and utilize separate backbones for background enhancement, we have conducted additional numerical comparison for resorting facial region only. In Table~\ref{tab:only_face}, we report the PSNR, SSIM, LPIPS, FVD, FID, and KID results for $8\times$ and $16\times$ video super-resolution on the short clips of CelebV-Text and CelebV-HQ datasets, respectively. As expected, our FLAIR quantitatively outperforms all other baseline methods in terms of both perception and data-fidelity metrics.

\noindent
\textbf{Effect of Data-Consistency.} We report PSNR and LPIPS results of our method in Fig.~\ref{fig:dcn} left and right for a mix of degradation consisting of $4\times$ SR, Gaussian blur and JPEG $Q=60$. We see that the perceptual quality (LPIPS) of the image improves as we use more number of iterations and remains after $K=40$. At the same time, the distortion (PSNR) drops accordingly, which is known as the trade-off between perception and distortion~\cite{Blau.etal2018perception}. More importantly, while both FLAIR w/ and w/o data-consistency module achieve similar LPIPS scores, FLAIR w/ data-consistency module better preserves the PSNR results.

\noindent
\textbf{Other Quantitative Results.} In Table~\ref{tab:x4_bicubic}, we report numerical results of FLAIR and some baseline methods for $4\times$ video super-resolution on two datasets. Note the better performance achieved by our FLAIR with different enhancement backbones even under mild degradation. In Table~\ref{tab:x4-stsr}, we report numerical results of using pre-trained FLAIR as spatial SR backbone for $4\times$ space-time video super-resolution. In Table~\ref{tab:obama}, we show quantative comparison of our FLAIR on Obama dataset for video motion deblurring. To further show that there is potential to adapt versatile backbones for our FLAIR enhancement module, we report numerical results of our FLAIR using the same pre-trained unconditional image DPM in~\eqref{eq:reverse4} as our enhancement backbone for $4\times$ SR, noisy Gaussian deblurring task. To demonstrate the adaptability of various backbones for our FLAIR enhancement module, we present numerical results where FLAIR employs the same pre-trained unconditional image DPM, as referenced in~\eqref{eq:reverse4}, as its enhancement backbone. For simplicity, we have limited our experiments to $4\times$ SR, noisy Gaussian deblurring task, deferring a more comprehensive evaluation to future work. The visual comparisons are shown in Fig~\ref{Fig:visua5564}. We make an interesting observation that FLAIR using unconditional image DPM as face enhancement module can improve the final restoration results in terms of PSNR and FVD on CelebV-Text. 

\noindent
\textbf{Evaluation of Running Time.} For completeness, we also report the running time of our FLAIR compared with the other image DPM baseline DDNM for $4\times$ SR video JPEG restoration in Table~\ref{tab:runtime}. It is worth to note that, while we observe that FLAIR exhibits relatively slow processing speeds, one may easily combine FLAIR with existing sampling acceleration methods, such as staring from refined $\xbm_K$~\cite{Chung.etal2022}, ODE based solvers~\cite{Lu.etal2022dpm, Liu.etal2022pseudo} and model distillation~\cite{Salimans.etal2022progressive}, etc.

\begin{table*}[]
\centering
\resizebox{\textwidth}{!}{%
\begin{tabular}{lccccc}
\hline
Hyperparameter                 & Bicubic 8 $\times$       & Bicubic 16 $\times$       & Gaussian Blur           & Motion Blur             & JPEG                    \\ \hline
\multicolumn{6}{c}{Model Architecture}                                                                                                             \\ \hline
Channels                       & 64               & 64               & 128                     & 128                     & 128                     \\
\# Resblocks                   & 1                & 1                & 2                       & 2                       & 2                       \\
Attention Resolutions          & (64, 32)         & (64, 32)         & (32, 16, 8)             & (32, 16, 8)             & (32, 16, 8)             \\
RFE Resolutions                & (512, 256)       & (512, 256)       & (512, 256)              & (512, 256)              & (512, 256)              \\
Channel Multiplier             & (1, 2, 4, 8, 16) & (1, 2, 4, 8, 16) & (0.5, 1, 1, 2, 2, 4, 4) & (0.5, 1, 1, 2, 2, 4, 4) & (0.5, 1, 1, 2, 2, 4, 4) \\
\# Attention Heads             & -                & -                & -                       & -                       & -                       \\
Head Channels                  & 64               & 64               & 64                      & 64                      & 64                      \\
Temporal Attention Window Size & 7                & 7                & 5                       & 5                       & 5                       \\ \hline
\multicolumn{6}{c}{Diffusion Setup}                                                                                                                \\ \hline
\# Diffusion Steps             & 2000             & 2000             & 1000                    & 1000                    & 1000                    \\
Noise Schedule                 & Linear           & Linear           & Linear                  & Linear                  & Linear                  \\
$\beta_1$                      & $1 \times 10^{-6}$             & $1 \times 10^{-6}$             & $1 \times 10^{-4}$                    & $1 \times 10^{-4}$                    & $1 \times 10^{-4}$                    \\
$\beta_T$                      & 0.01             & 0.01             & 0.02                    & 0.02                    & 0.02                   \\ \hline
\multicolumn{6}{c}{Image DPM Training}                                                                                                                 \\ \hline
Batch size                     & 64               & 64               & 64                      & 64                      & 64                      \\
Learning Rate                  & $1.5 \times 10^{-4}$             & $1.5 \times 10^{-4}$             & $1.5 \times 10^{-4}$                    & $1.5 \times 10^{-4}$                    & $1.5 \times 10^{-4}$                    \\
Weight Decay                   & 0.05             & 0.05             & 0.05                    & 0.05                    & 0.05                    \\
\# Samples                     & 2M               & 2M               & 2M                      & 2M                      & 2M                      \\
EMA rate                       & 0.9999           & 0.9999           & 0.9999                  & 0.9999                  & 0.9999                  \\ \hline
\multicolumn{6}{c}{Video DPM Fine-tuning}                                                                                                              \\ \hline
Batch size                     & 4                & 4                & 4                       & 4                       & 4                      \\
Frame Length $N$                     & 10                & 10                & 10                       & 10                       & 10                      \\
Learning Rate                  & $1 \times 10^{-4}$             & $1 \times 10^{-4}$             & $1 \times 10^{-4}$                    & $1 \times 10^{-4}$                    & $1 \times 10^{-4}$                    \\
Weight Decay                   & 0.05             & 0.05             & 0.05                    & 0.05                    & 0.05                    \\
\# Samples                     & 0.3M             & 0.3M             & 0.3M                    & 0.3M                    & 0.3M                    \\
EMA rate                       & -                & -                & -                       & -                       & -                       \\ \hline
\multicolumn{6}{c}{Sampling}                                                                                                                       \\ \hline
$\|K\|$                        & 25               & 100              & 100                     & 65                      & 40                      \\
$\rho_t$                         & 0.85             & 0.85             & 0.25                    & 0.35                    & 0.5                     \\
$w_\tau$                       & 0.85             & 0.7              & 0.75                    & 0.1                     & 0.5                     \\
$\tau$                         & 5                & 5                & 5                       & 5                       & 5                       \\
$\zeta$                        & -                & -                & 1000                    & 1000                    & 1000                    \\ \hline
\end{tabular}%
}
\caption{Hyperparameters used in our FLAIR implementations.}
\label{tab:hparam}
\end{table*}

\subsection{Additional Visual Results} 
\begin{figure*}[t!]
	\centering
	\includegraphics[width=0.80\textwidth]{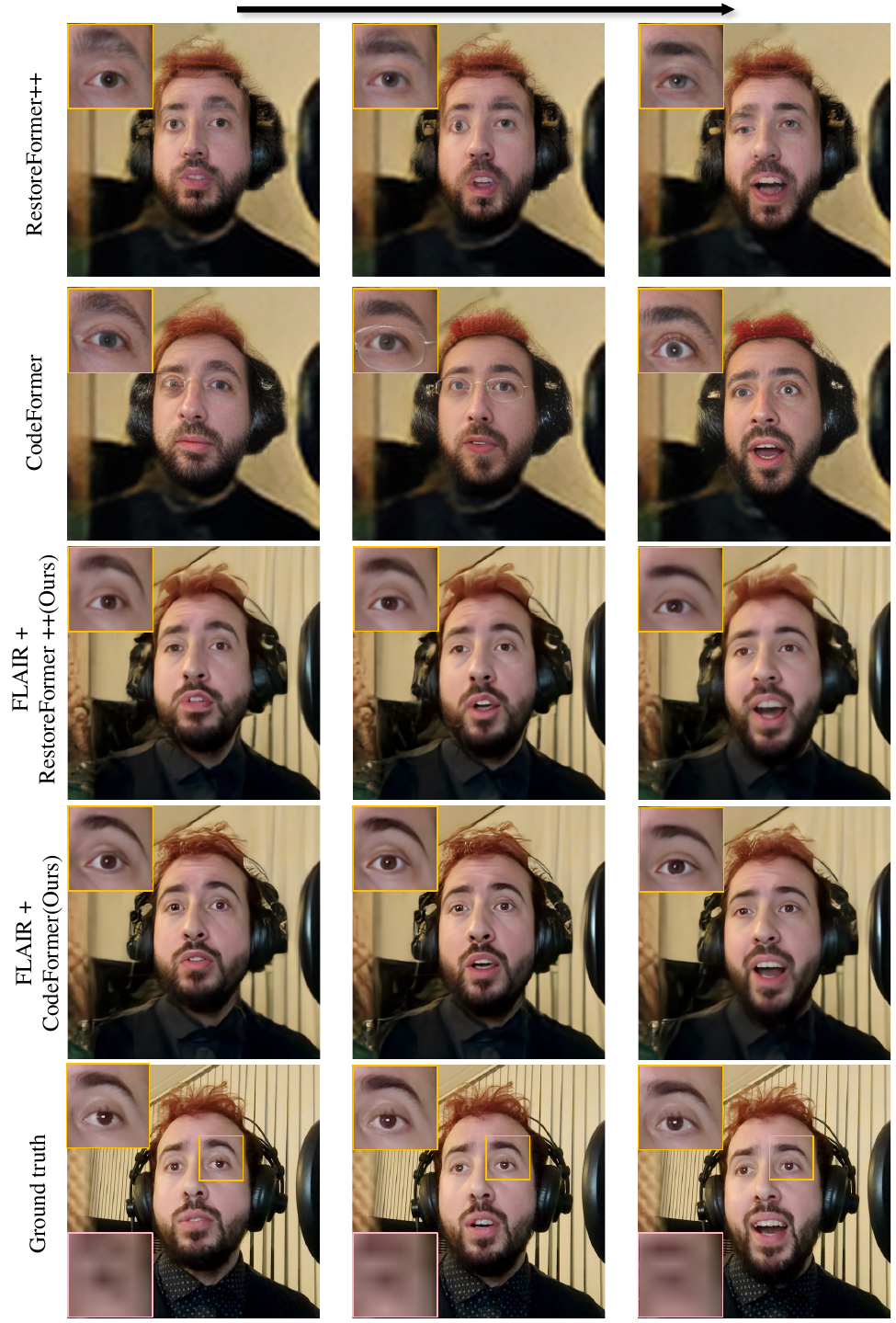}
    \vspace{-1em}
	\caption{More visual results of $16\times$ video super-resolution on CelebV-Text~\cite{Yu2023.etalcelebv} dataset. Each row consists of three video frames, with an interval of five frames between each selected frame. The zoomed-in regions of each method are displayed in yellow boxes, along with their LQ counterparts in pink boxes. Best viewed by zooming in the display.}
	\label{Fig:visual0}
\vspace{-0.em}
\end{figure*}

\begin{figure*}[t!]
	\centering
	\includegraphics[width=0.80\textwidth]{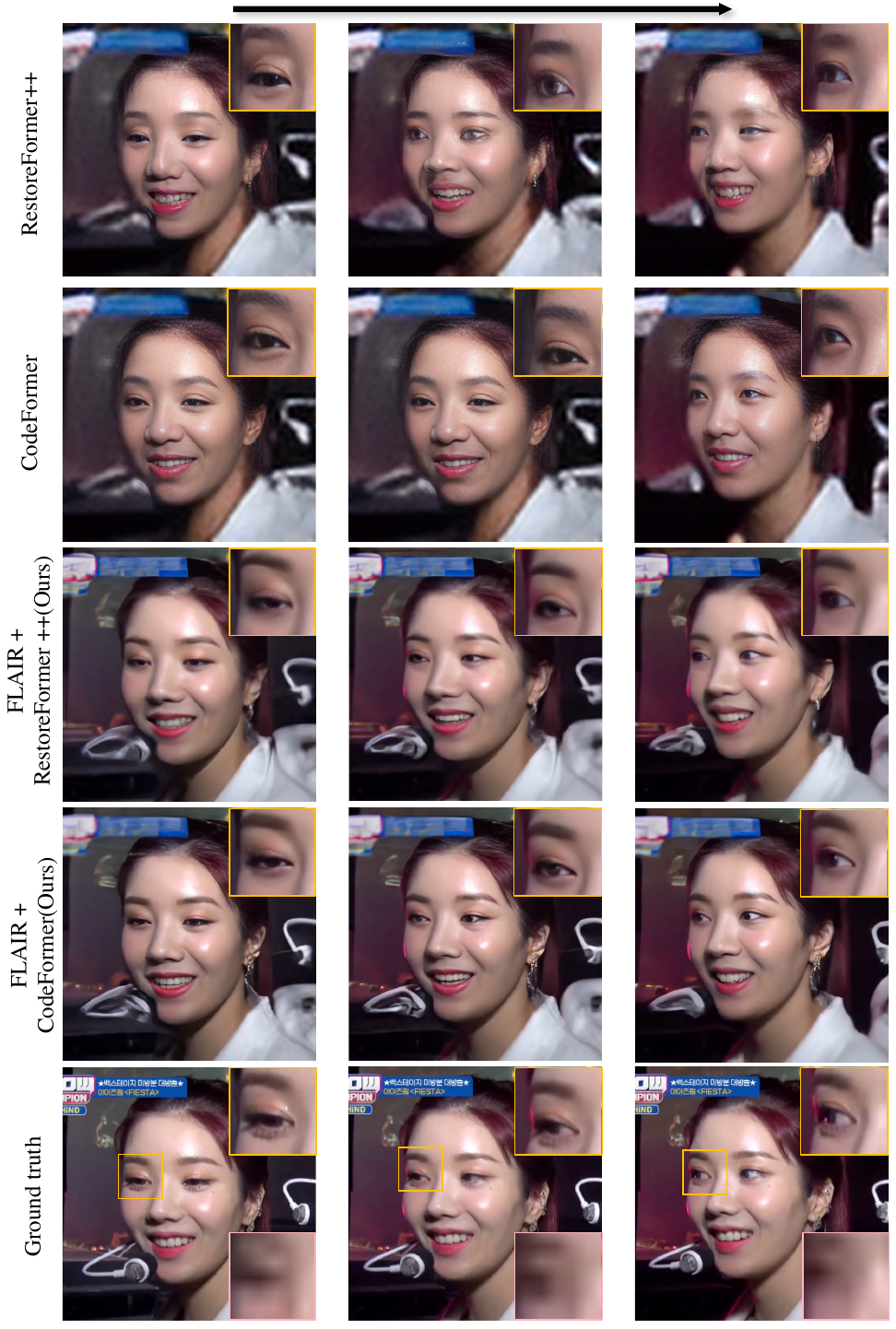}
    \vspace{-1em}
	\caption{More visual results of $16\times$ video super-resolution on CelebV-Text~\cite{Yu2023.etalcelebv} dataset. Each row consists of three video frames, with an interval of five frames between each selected frame. The zoomed-in regions of each method are displayed in yellow boxes, along with their LQ counterparts in pink boxes. Best viewed by zooming in the display.}
	\label{Fig:visual2983}
\vspace{-0.em}
\end{figure*}

\begin{figure*}[t!]
	\centering
	\includegraphics[width=0.80\textwidth]{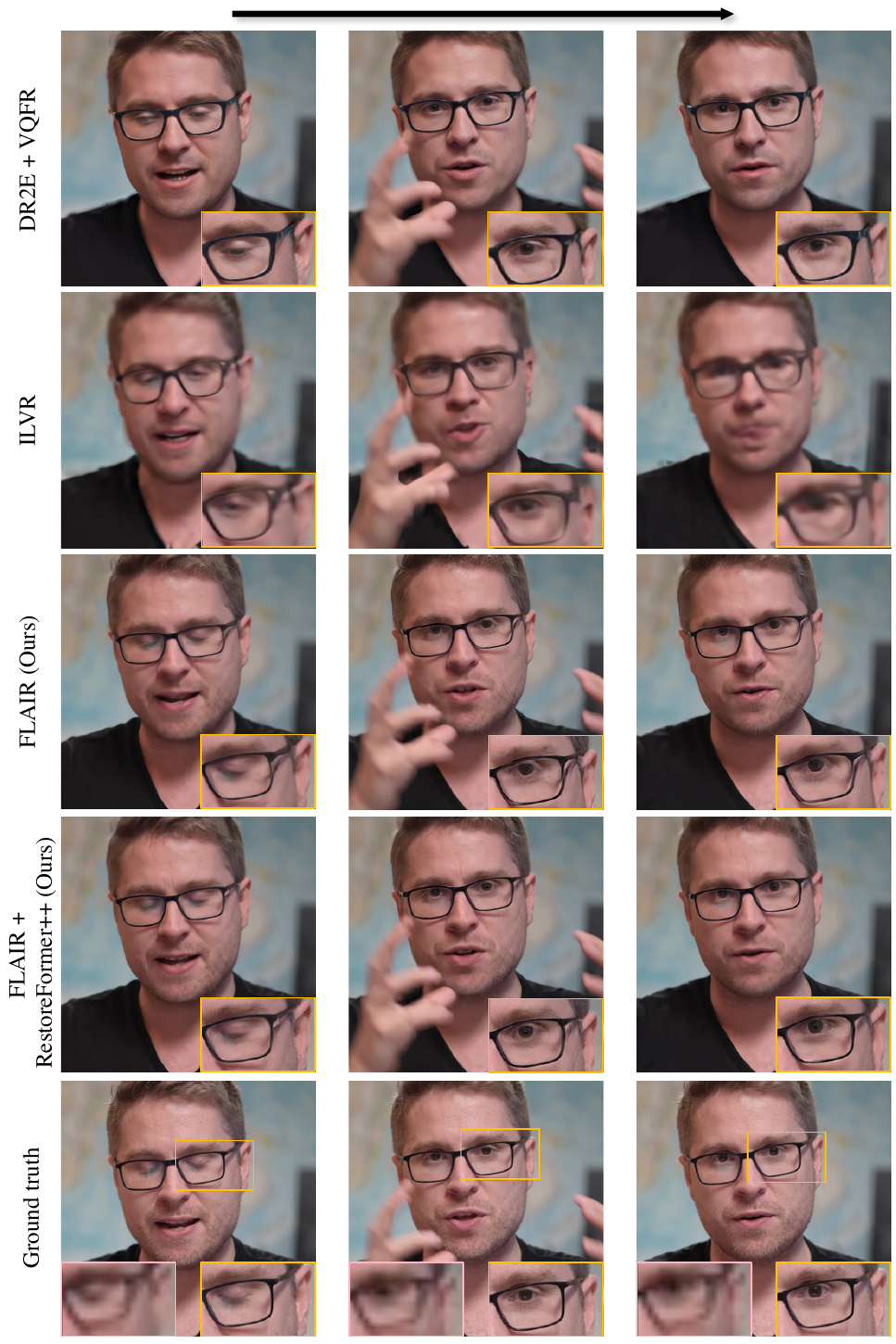}
    \vspace{-0.5em}
\caption{More visual comparisons of $8\times$ face video super-resolution on CelebV-Text~\cite{Yu2023.etalcelebv}. Each row consists of three video frames, with an interval of five frames between each selected frame. The zoomed-in regions of each method are displayed in yellow boxes, along with their LQ counterparts in pink boxes. Best viewed by zooming in the display.}
\label{Fig:visua30}
\vspace{-0.em}
\end{figure*}

\begin{figure*}[t!]
	\centering
	\includegraphics[width=0.80\textwidth]{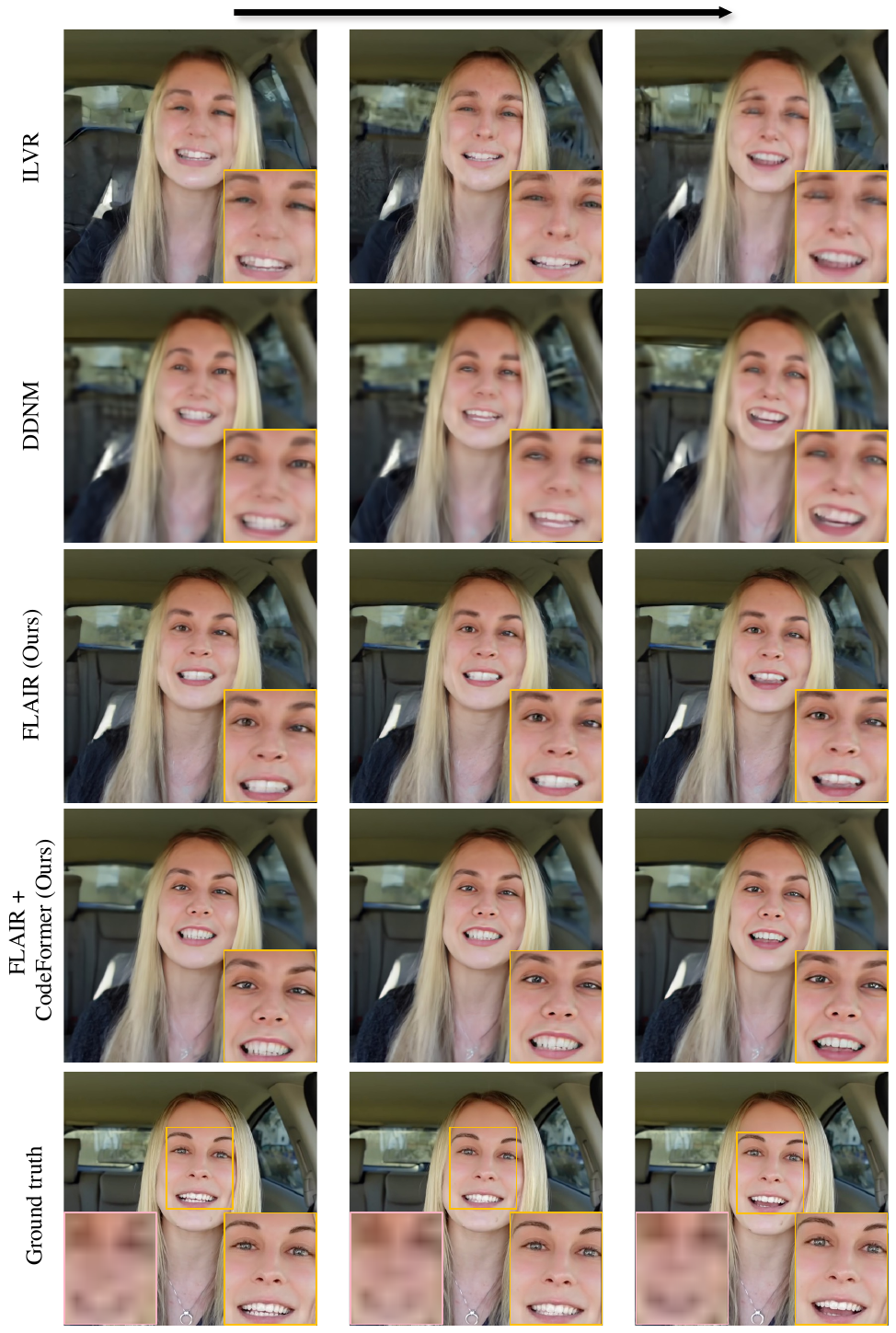}
    \vspace{-0.5em}
\caption{More visual comparisons of $16\times$ face video super-resolution on CelebV-HQ~\cite{Zhu.etal2022celebv}. Each row consists of three video frames, with an interval of five frames between each selected frame. The zoomed-in regions of each method are displayed in yellow boxes, along with their LQ counterparts in pink boxes. Best viewed by zooming in the display.}
	\label{Fig:visual7}
\vspace{-0.em}
\end{figure*}

\begin{figure*}[t!]
	\centering
	\includegraphics[width=0.80\textwidth]{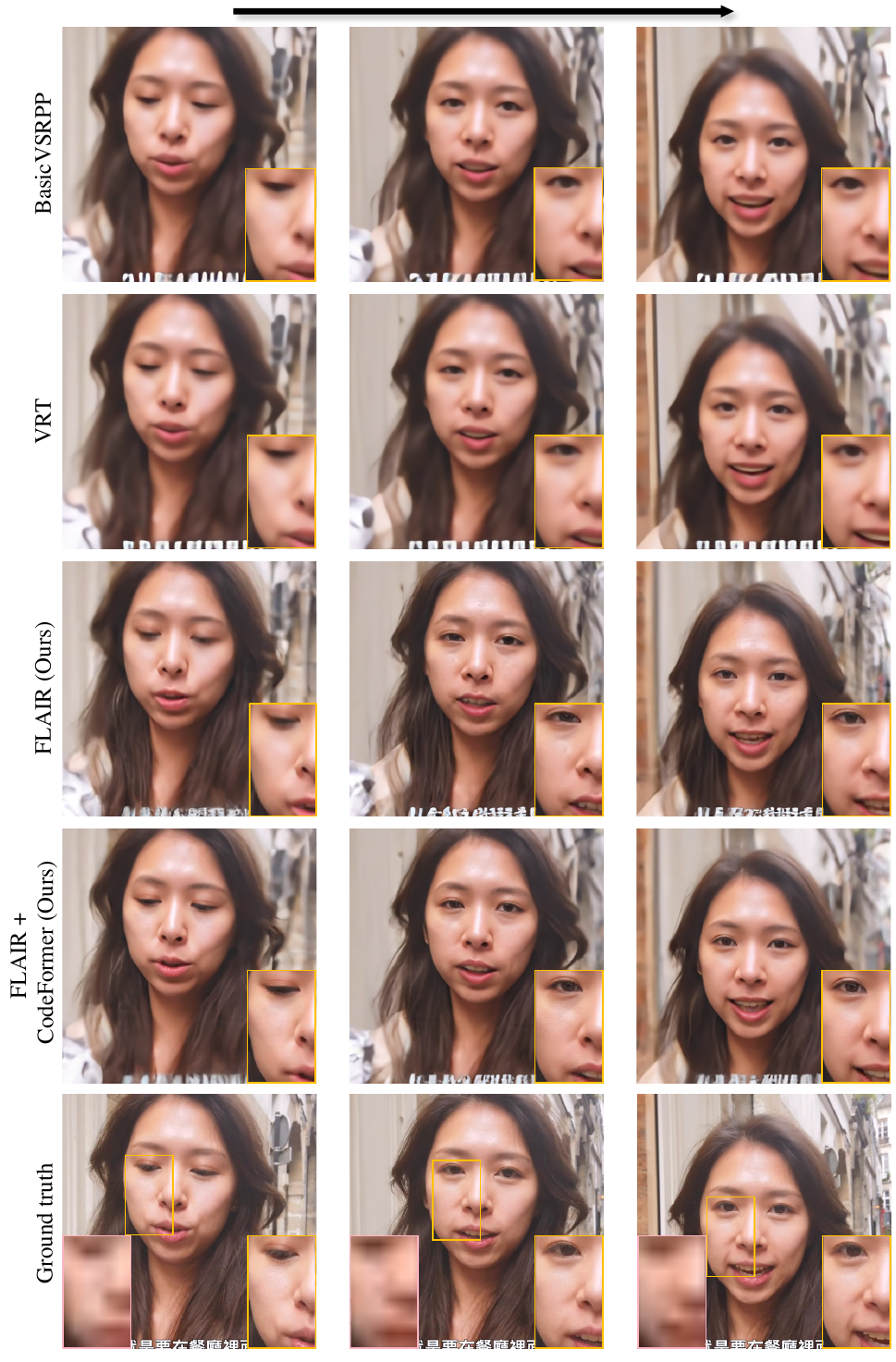}
    \vspace{-1em}
	\caption{More visual comparisons of $16\times$ face video super-resolution on CelebV-Text~\cite{Yu2023.etalcelebv}. Each row consists of three video frames, with an interval of five frames between each selected frame. The zoomed-in regions of each method are displayed in yellow boxes, along with their LQ counterparts in pink boxes. Best viewed by zooming in the display.}
	\label{Fig:visual2}
\vspace{-0.em}
\end{figure*}

\begin{figure*}[t!]
	\centering
	\includegraphics[width=0.80\textwidth]{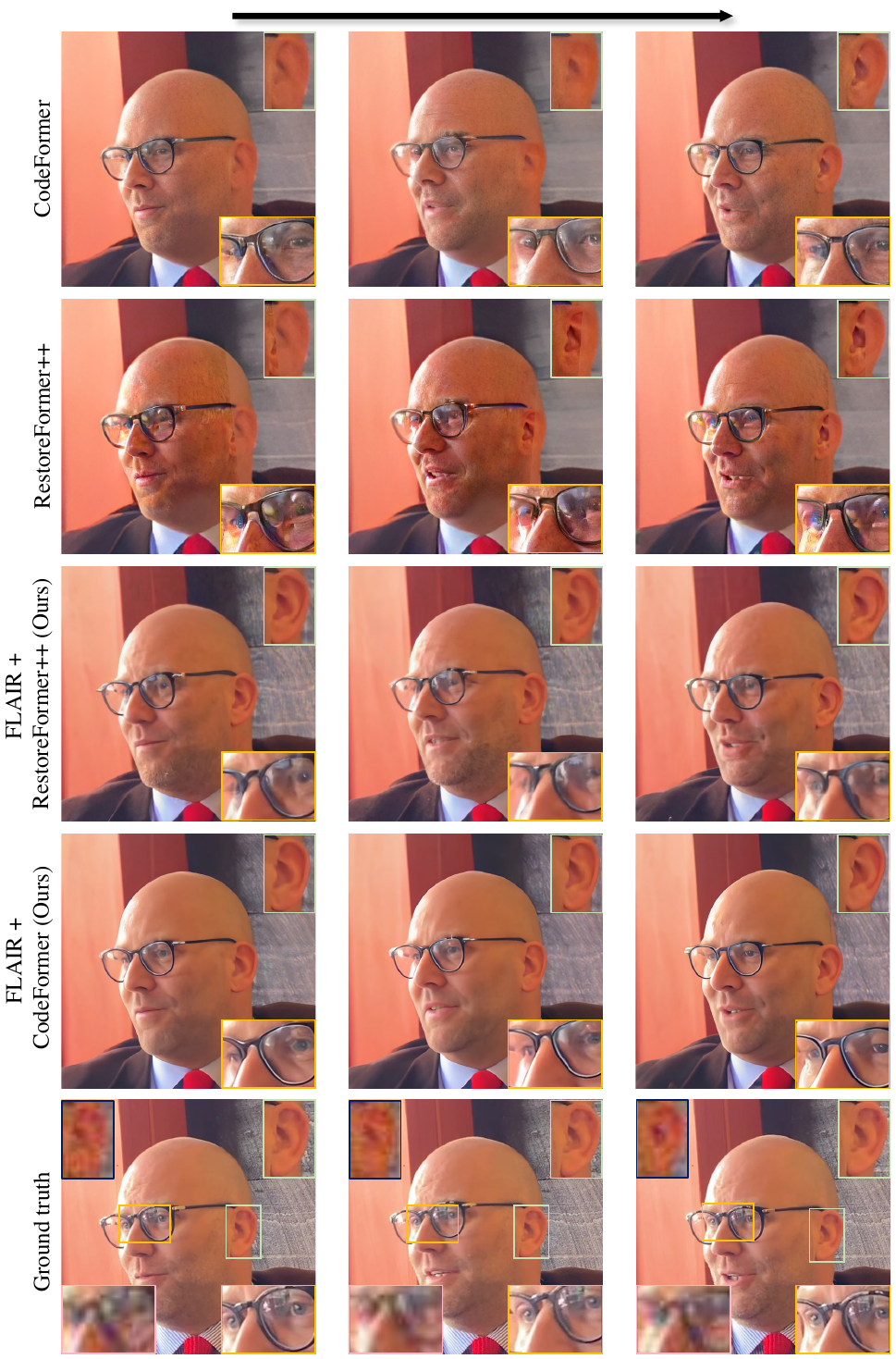}
    \vspace{-0.5em}
\caption{More visual comparisons of $4\times$ face video JPEG restoration on CelebV-Text~\cite{Yu2023.etalcelebv} dataset. Each row consists of three video frames, with an interval of five frames between each selected frame. The zoomed-in regions of each method are displayed in yellow and green boxes, along with their LQ counterparts in pink and blue boxes. Best viewed by zooming in the display.}
	\label{Fig:visua7564}
\vspace{-0.em}
\end{figure*}

\begin{figure*}[t!]
	\centering
	\includegraphics[width=0.80\textwidth]{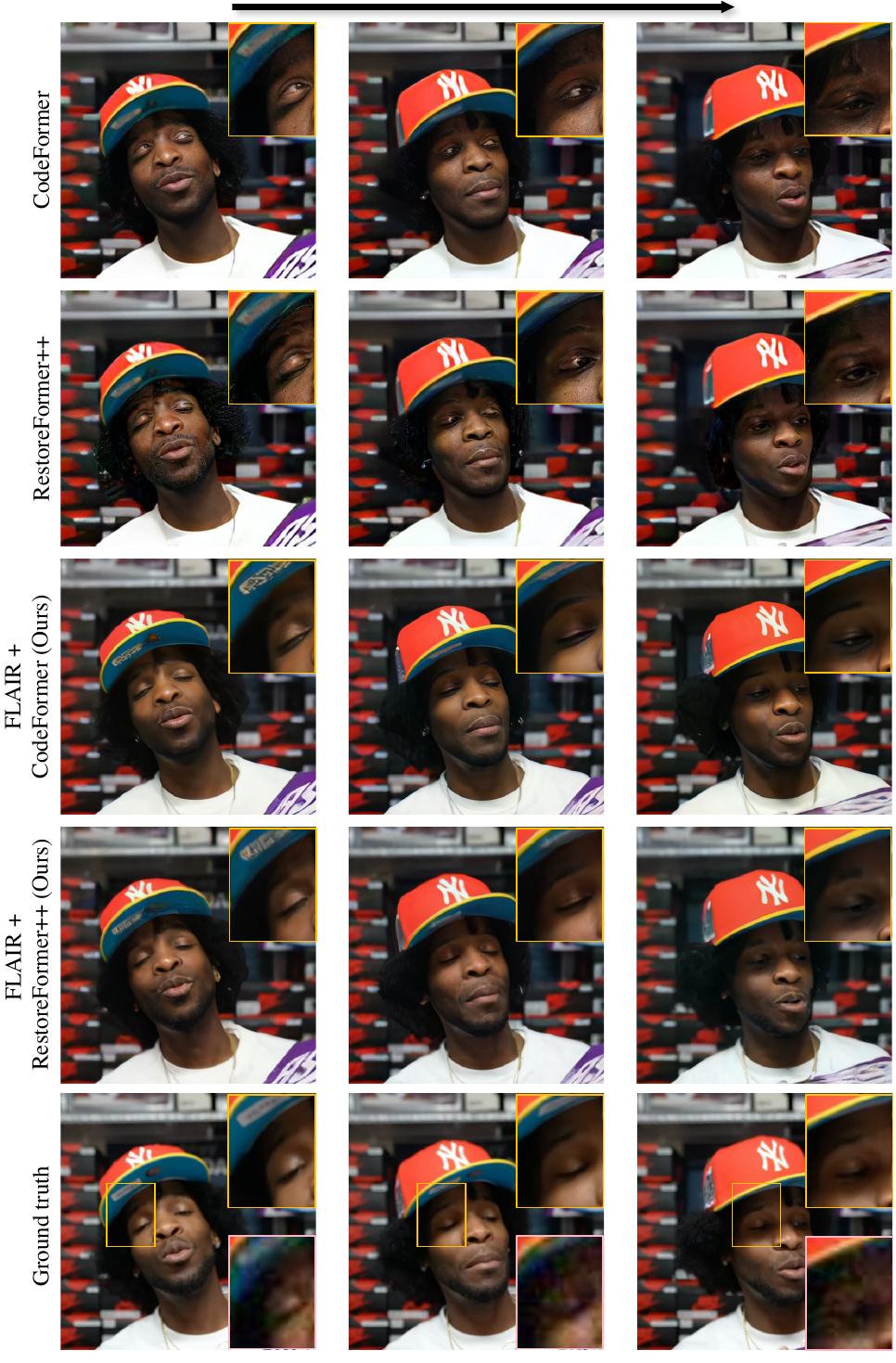}
    \vspace{-0.5em}
\caption{Visual comparisons of $4\times$ face video JPEG restoration on CelebV-Text~\cite{Yu2023.etalcelebv}. Each row consists of three video frames, with an interval of five frames between each selected frame. The zoomed-in regions of each method are displayed in yellow boxes, along with their LQ counterparts in pink boxes. Best viewed by zooming in the display.}
	\label{Fig:visua2235}
\vspace{-0.em}
\end{figure*}

\begin{figure*}[t!]
	\centering
	\includegraphics[width=0.80\textwidth]{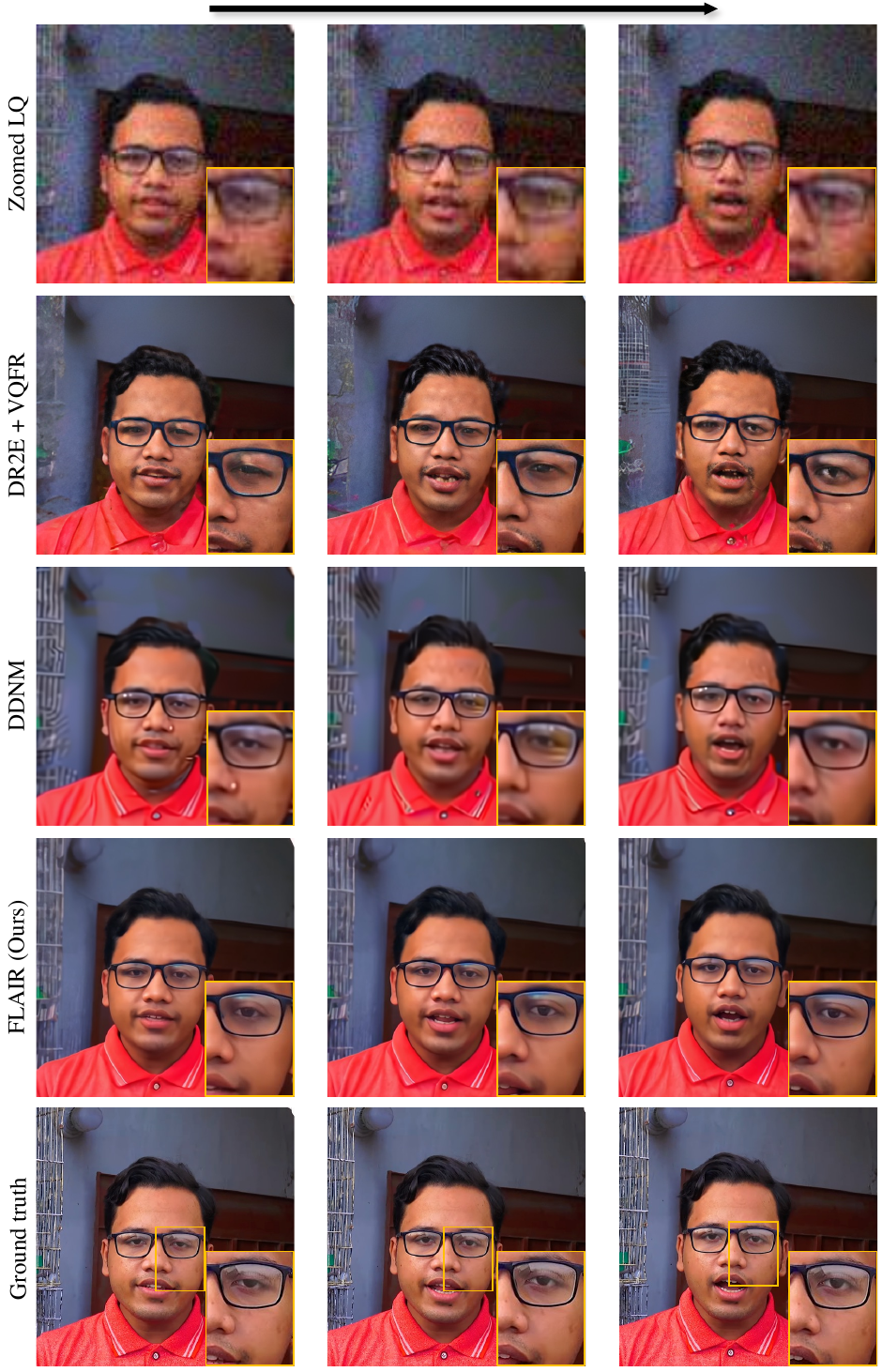}
    \vspace{-0.5em}
\caption{Visual comparisons of $4\times$ face video JPEG restoration on CelebV-HQ~\cite{Zhu.etal2022celebv}. Each row consists of three video frames, with an interval of five frames between each selected frame. The zoomed-in regions of each method are displayed in yellow boxes. Best viewed by zooming in the display.}
	\label{Fig:visua6235}
\vspace{-0.em}
\end{figure*}

\begin{figure*}[t!]
	\centering
	\includegraphics[width=0.80\textwidth]{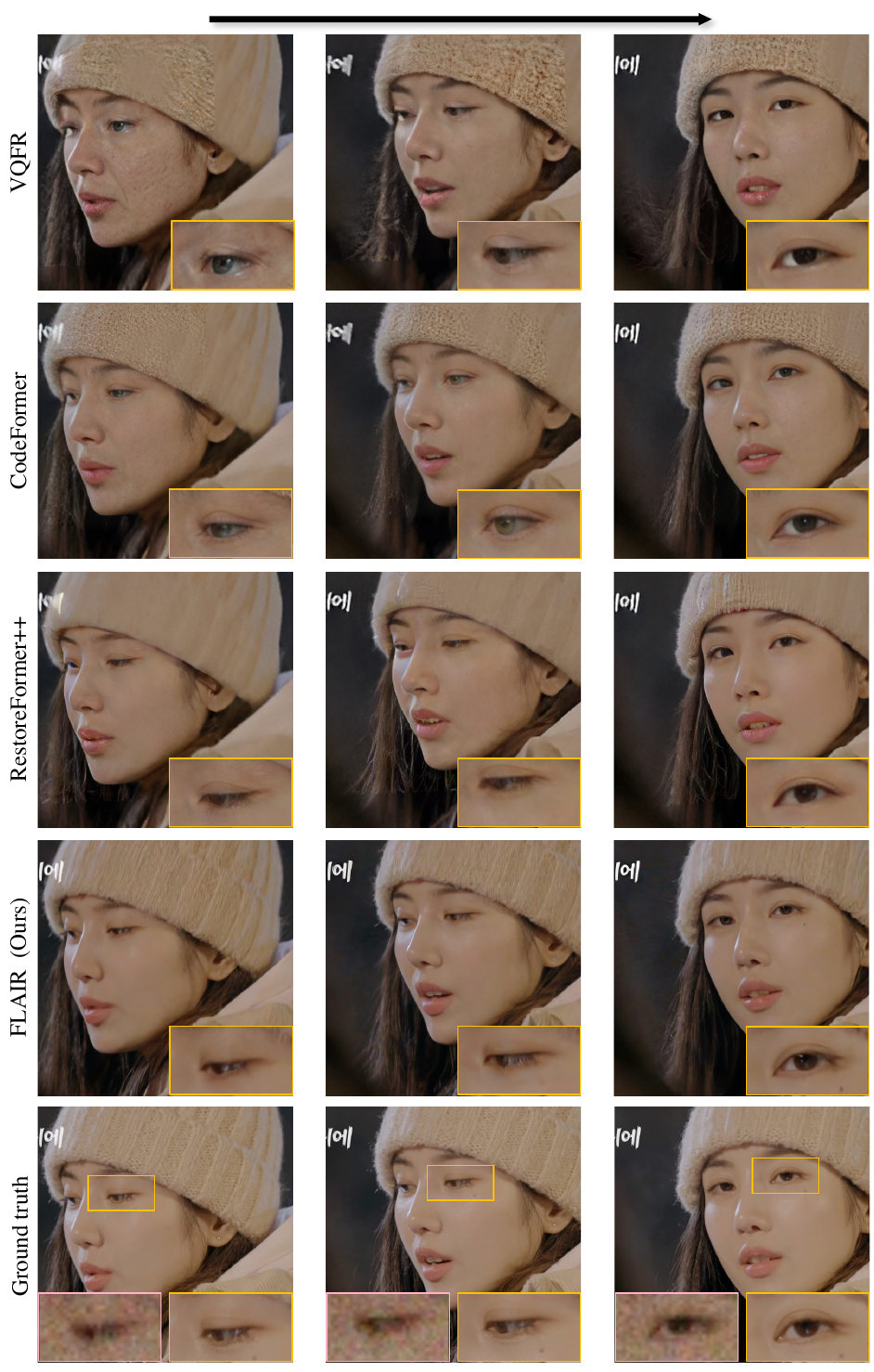}
    \vspace{-0.5em}
\caption{Visual comparisons of $4\times$ face video motion deblurring on CelebV-Text~\cite{Yu2023.etalcelebv}. Each row consists of three video frames, with an interval of ten frames between each selected frame. The zoomed-in regions of each method are displayed in yellow boxes, along with their LQ counterparts in pink boxes. Best viewed by zooming in the display.}
	\label{Fig:visua65}
\vspace{-0.em}
\end{figure*}

\begin{figure*}[t!]
	\centering
	\includegraphics[width=0.80\textwidth]{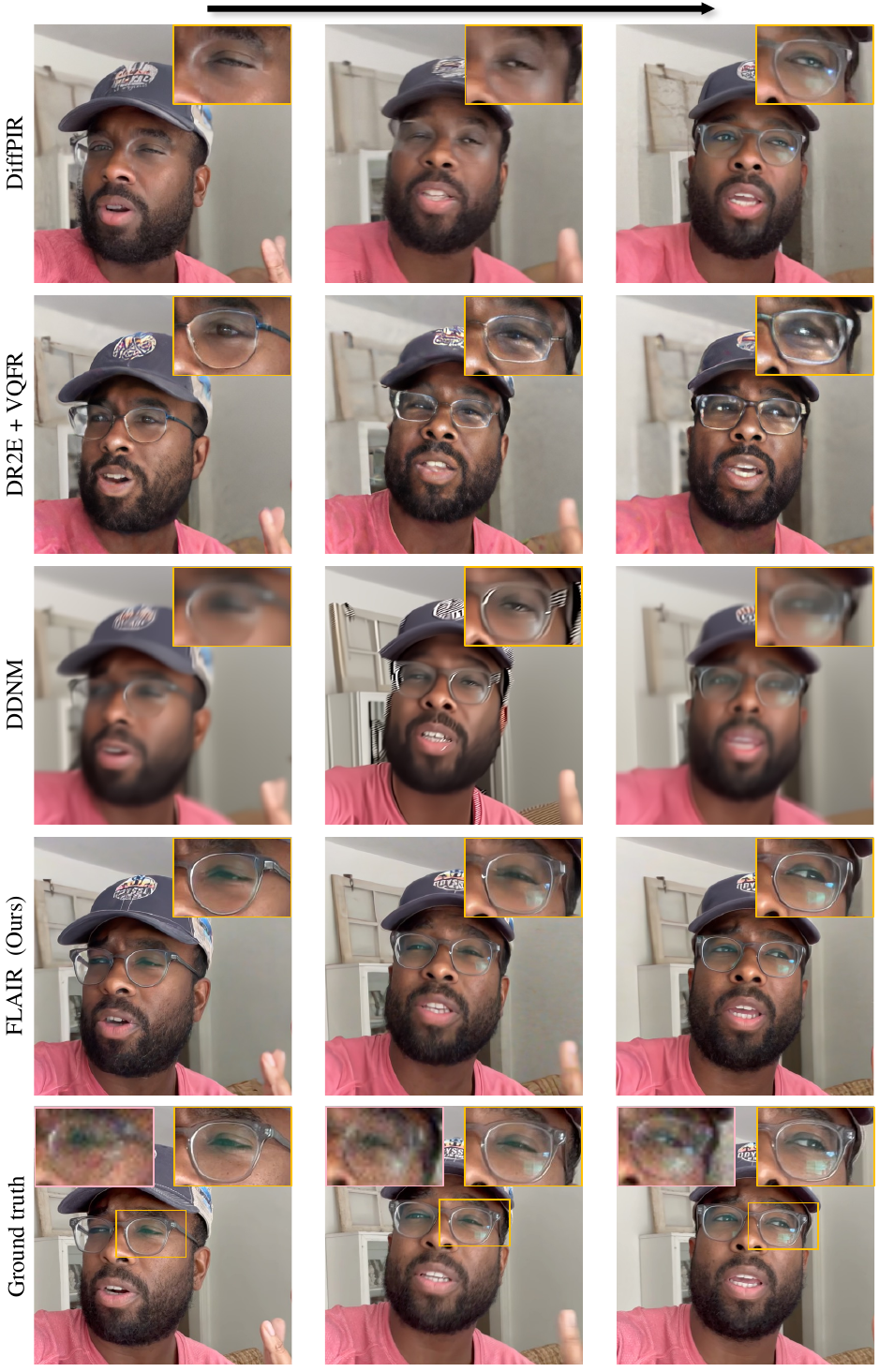}
    \vspace{-0.5em}
\caption{Visual comparisons of $4\times$ face video motion deblurring on CelebV-Text~\cite{Yu2023.etalcelebv}. Each row consists of three video frames, with an interval of ten frames between each selected frame. The zoomed-in regions of each method are displayed in yellow boxes, along with their LQ counterparts in pink boxes. Best viewed by zooming in the display.}
	\label{Fig:visua50}
\vspace{-0.em}
\end{figure*}

\begin{figure*}[t!]
	\centering
	\includegraphics[width=0.80\textwidth]{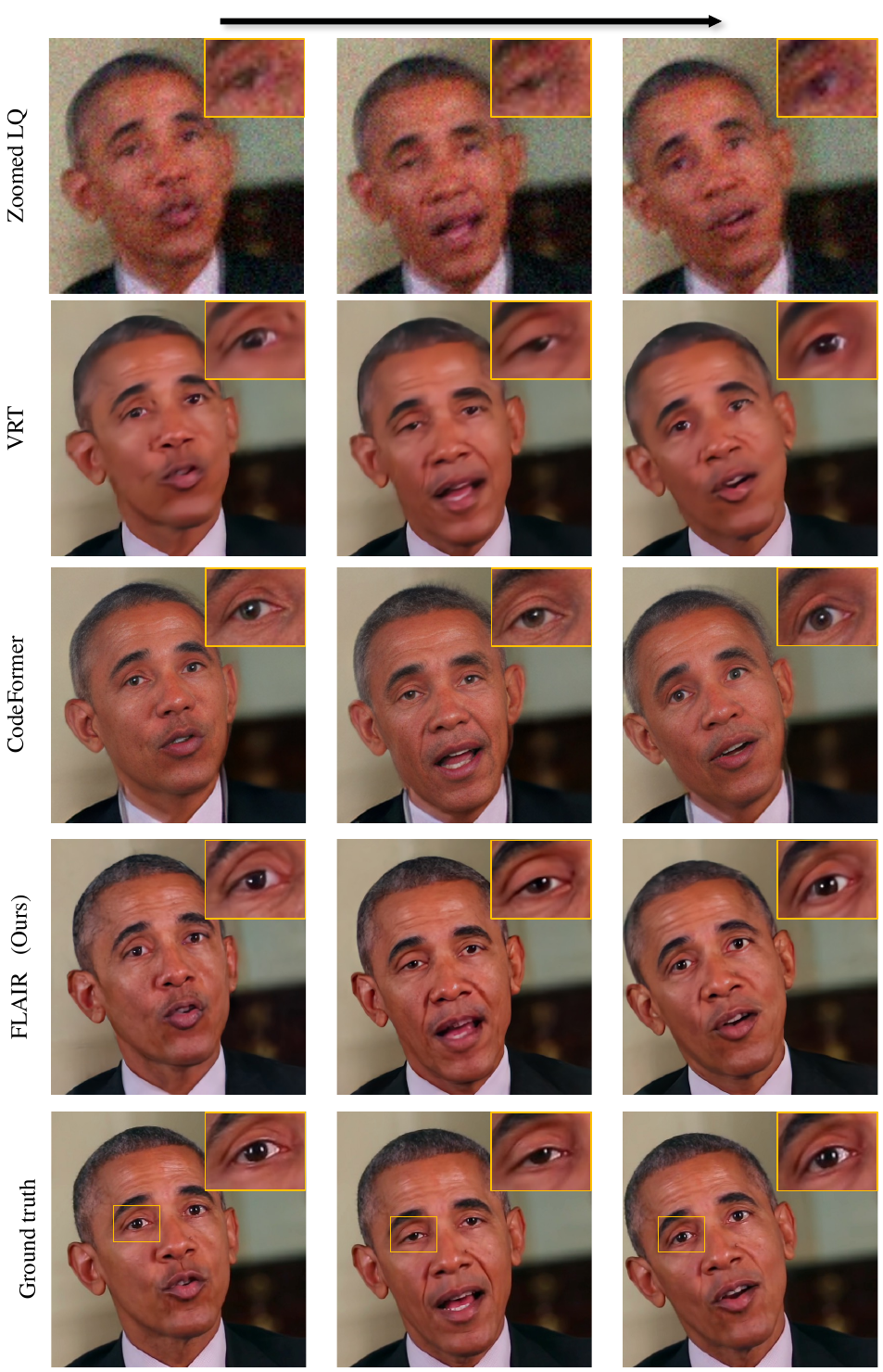}
    \vspace{-0.5em}
\caption{Visual comparisons of $4\times$ face video motion deblurring on Obama dataset~\cite{Suwajanakorn.etal2017synthesizing}. Each row consists of three video frames, with an interval of five frames between each selected frame. The zoomed-in regions of each method are displayed in yellow boxes. Best viewed by zooming in the display.}
	\label{Fig:visua7852}
\vspace{-0.em}
\end{figure*}

\begin{figure*}[t!]
	\centering
	\includegraphics[width=0.92\textwidth]{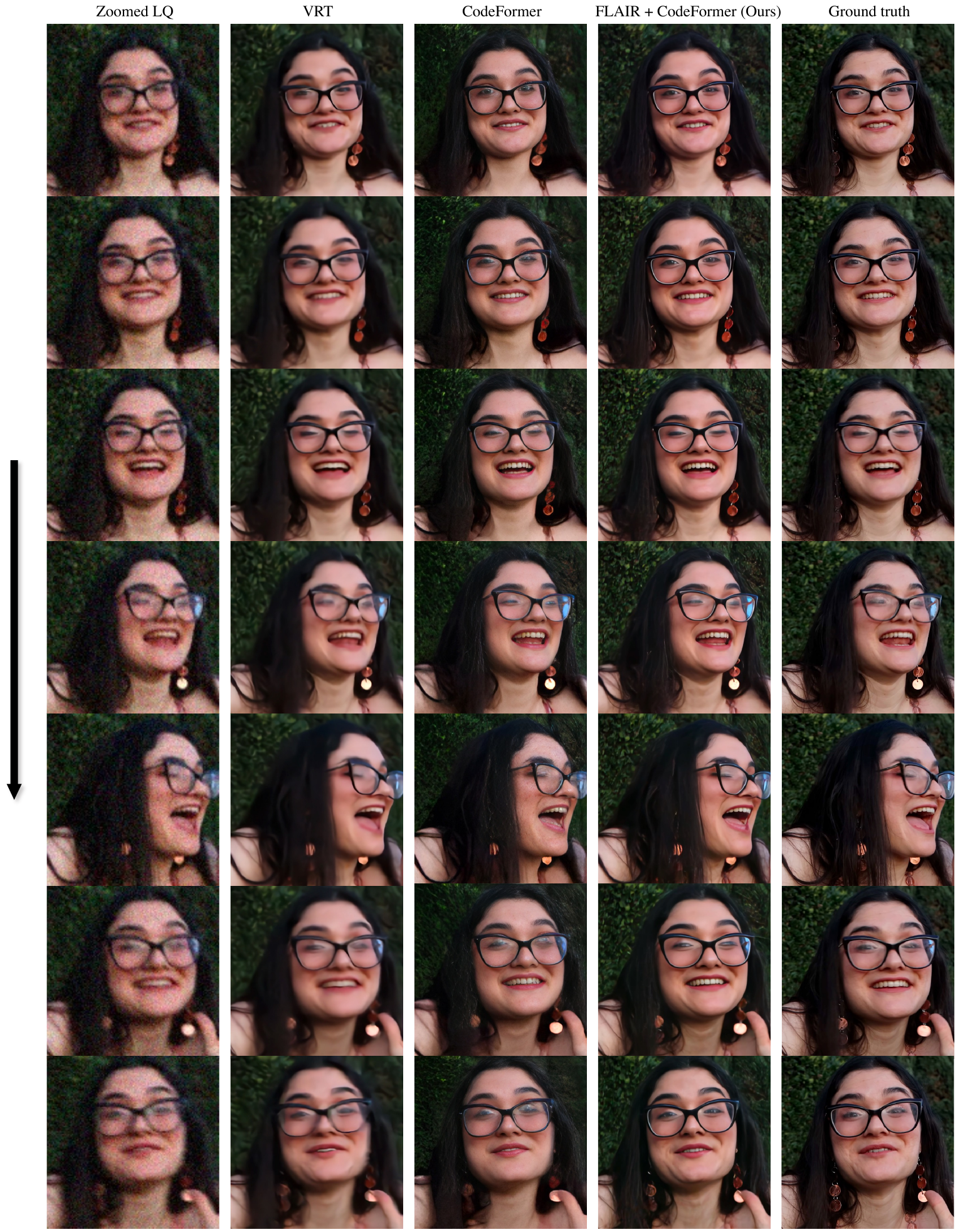}
    \vspace{-1em}
	\caption{Visual comparisons of $4\times$ face video motion deblurring on CelebV-Text~\cite{Yu2023.etalcelebv}. Each column consists of seven video frames, with an interval of ten frames between each selected frame. Best viewed by zooming in the display.}
	\label{Fig:visual1} 
\vspace{-0.em}
\end{figure*}

\begin{figure*}[t!]
	\centering
	\includegraphics[width=0.94\textwidth]{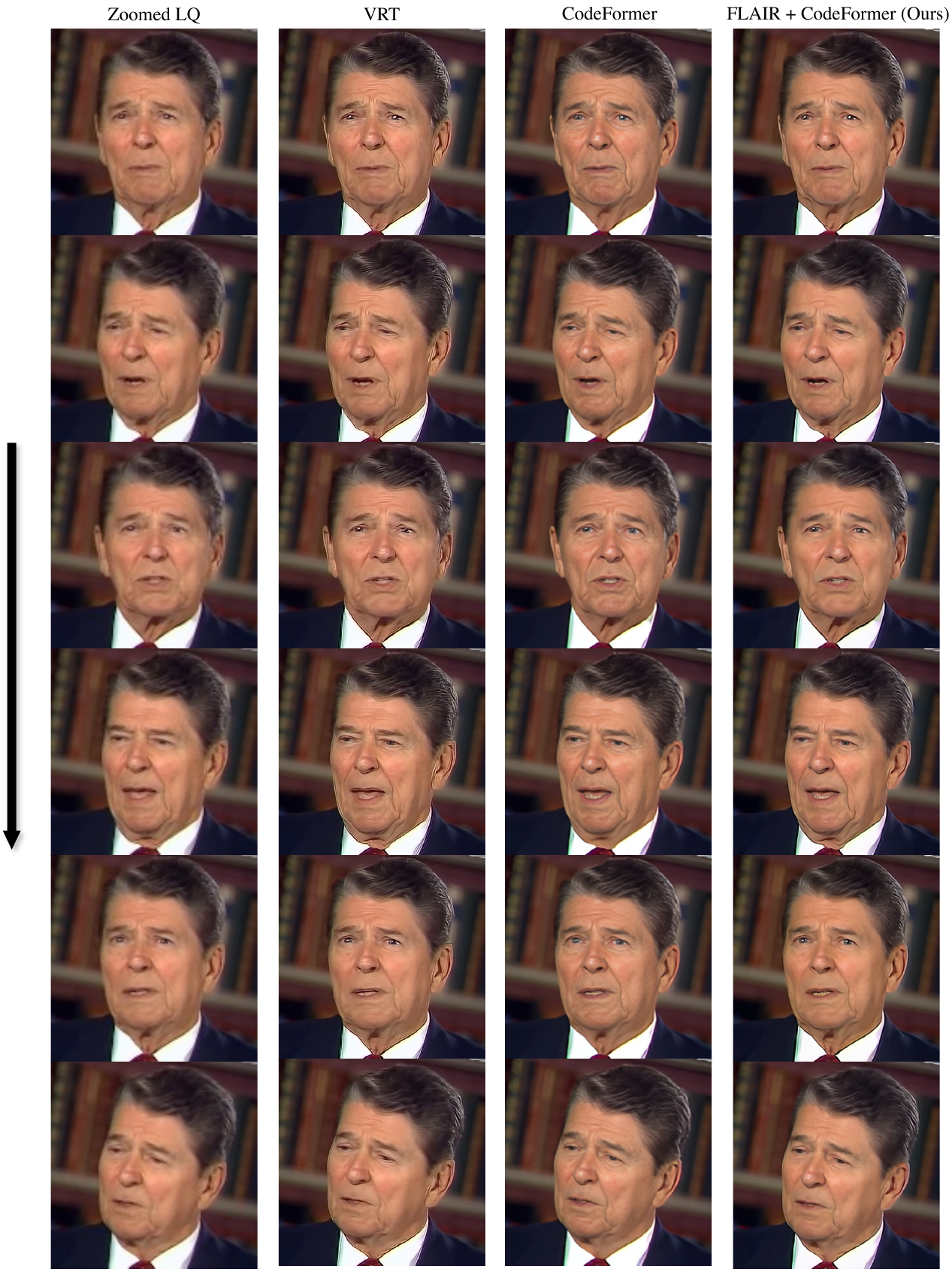}
    \vspace{-1em}
	\caption{Visual comparisons of~\emph{real-world} web video enhancement. Each column consists of six video frames, with an interval of around fifteen frames between each selected frame. Best viewed by zooming in the display.}
	\label{Fig:visual15}
\vspace{-0.em}
\end{figure*}

In Figs.~\ref{Fig:visual0} - \ref{Fig:visual2}, we present additional visual comparisons of several methods for video super-resolution on CelebV-Text and CelebV-HQ, where each row contains three frames. For each case, we also provide the zoomed-in region of the degraded inputs accordingly. In Figs.~\ref{Fig:visua7564} - \ref{Fig:visua6235}, we show more visual comparisons of several methods for video JPEG restoration with the zoomed-in regions. For video deblurring, we present the visual results through Fig.\ref{Fig:visua65} to \ref{Fig:visual1}. For real-world web video enhancement task, we assume the LQ inputs $\ybm$ corrupted by mixed degradation. Since our video DPM is trained for multi-variant degradation, we only need to fine-tune the data-consistency module. By fine-tuning the forward-model such that $\Acal\Acal^{+}\ybm \approx \ybm$, we observe that the degradation of $4\times$ SR with Gaussian kernel of width$=1.6$, JPEG $Q=90$ works the best. In Fig.~\ref{Fig:visual15}, we present more visual results of our FLAIR compared with several baseline methods. One can see
from Fig.~\ref{Fig:visual15} that our designed two stage enhancement modules together can improve visual quality while preserving the data-consistency effectively.

\newpage
\newpage
\newpage
\newpage
\newpage
\newpage
\newpage
\newpage
\newpage
\newpage
\newpage
\newpage

\newpage

\end{document}